\documentclass[10pt,twocolumn,letterpaper]{article}

\usepackage{iccv}
\usepackage{times}
\usepackage{epsfig}
\usepackage{graphicx}
\usepackage{amssymb,amsthm,amsmath}

\usepackage{microtype}
\usepackage{subfigure}
\usepackage{booktabs} 

\usepackage{url}            
\usepackage{tabularx}       
\usepackage{multirow}
\usepackage{amsfonts}       
\usepackage{nicefrac}       
\usepackage{epigraph}
\usepackage{soul}
\usepackage[dvipsnames]{xcolor}
\usepackage{paralist}
\usepackage{color, colortbl}

\usepackage{pifont} 
\usepackage{wrapfig}

\usepackage{makecell}
\usepackage{array}
\usepackage[export]{adjustbox}

\usepackage{algorithm}
\usepackage{listings}
\usepackage{xcolor}

\usepackage{framed}
\usepackage{mathtools}
\usepackage{verbatim} 
\usepackage{enumerate}
\usepackage{bbm}
\usepackage{amsbsy}
\usepackage{float}
\usepackage{enumitem}
\usepackage[utf8x]{inputenc} 
\usepackage[T1]{fontenc}    
\usepackage{bm}
\usepackage{textcomp}
\usepackage{setspace}

\usepackage[toc,page]{appendix}
\usepackage[margin=0.9in,headsep=-20pt]{geometry}
\usepackage{mathtools, cuted}
\usepackage{algorithmic}

\newcommand{\cit}[1]{\cite{#1}}

\newcommand{\loss}{CrossCLR }

\definecolor{mygreen}{rgb}{0.032, 0.6392, 0.2039}
\newcommand{\cmark}{\textcolor{mygreen}{\ding{51}}}%
\newcommand{\xmark}{\textcolor{red}{\ding{55}}}%

\newcommand{\std}[1]{\tiny{$\pm$#1}}

\newcommand{\tbf}[1]{\textbf{#1}}

\newcommand{\thickhline}{
    \specialrule{.1em}{0em}{0em}
}

\newcommand{\ignore}[1]{} 

\graphicspath{ {./figures/} }

\usepackage[pagebackref=true,breaklinks=true,letterpaper=true,colorlinks,bookmarks=false]{hyperref}

 \iccvfinalcopy 

\def\iccvPaperID{10625} 

\begin{document}

\title{\large CrossCLR: Cross-modal Contrastive Learning For Multi-modal Video Representations}

\author{Mohammadreza Zolfaghari$^{1*}$, Yi Zhu$^{2}$, Peter Gehler$^{2}$, Thomas Brox$^{2}$\\
$^{1}$University of Freiburg\hspace{6pt} $^2$Amazon\hspace{6pt}}

\let\svthefootnote\thefootnote
\newcommand\freefootnote[1]{%
  \let\thefootnote\relax%
  \footnotetext{#1}%
  \let\thefootnote\svthefootnote%
}

\maketitle

\begin{abstract}
Contrastive learning allows us to flexibly define powerful losses by contrasting positive pairs from sets of negative samples. 
Recently, the principle has also been used to learn cross-modal embeddings for video and text, yet without exploiting its full potential. 
In particular, previous losses do not take the intra-modality similarities into account, which leads to inefficient embeddings, as the same content is mapped to multiple points in the embedding space. 
With CrossCLR, we present a contrastive loss that fixes this issue.
Moreover, we define sets of highly related samples in terms of their input embeddings and exclude them from the negative samples to avoid issues with false negatives. 
We show that these principles consistently improve the quality of the learned embeddings. 
The joint embeddings learned with CrossCLR extend the state of the art in video-text retrieval on Youcook2 and LSMDC datasets and in video captioning on Youcook2 dataset by a large margin.  
We also demonstrate the generality of the concept by learning improved joint embeddings for other pairs of modalities.  
\end{abstract}

\freefootnote{$^*$ Work done during an internship at Amazon Tübingen.}

\section{Introduction}
\label{submission}
%



Cross-modal tasks, especially tasks connecting video and text, expand the influence and applicability of computer vision. It enables, for example, video retrieval based on text queries~\cite{gabeur2020multimodal, DongLXJH0W19, LiuANZ19}, image and video captioning~\cite{ging2020coot}, and exploitation of text-based meta-data for visual feature learning~\cite{miech19endtoend, miech19howto100m, actbert20,9009570}.  
Linking the different, not directly comparable sources of data creates new challenges that do not appear in visual-only learning.  

In this paper, we consider cross-modal contrastive learning and introduce a loss that relates the data in a more efficient manner than direct adoption of a loss designed for vision-only data. 
Contrastive learning is based on the definition of positive and negative samples relative to an ankor point, which yields a flexible principle: pull together an anchor and a positive sample in the embedding space, and push apart the anchor from many negative samples.
Many implementations of this principle have been proposed: max-margin loss~\cite{1640964}, triplet loss~\cite{triplet1, triplet2, triplet3}, and InfoNCE~\cite{abs-1807-03748}. 
Usually positive pairs are defined as synthetic, spatial~\cite{chen_icml2020_simclr, rui2020}, or temporal variations of an instance~\cite{rui2020}. 
Instance discrimination has been applied  also to cross-modal tasks, where positive pairs (or a set of positives) are sampled from the same time window (MILNCE~\cite{miech19endtoend}, AVSA~\cite{morgado2020learning}, CM-ACC~\cite{ma2020learning}).

\begin{figure}[t]
    \begin{center}
    \centerline{\includegraphics[width=\columnwidth]{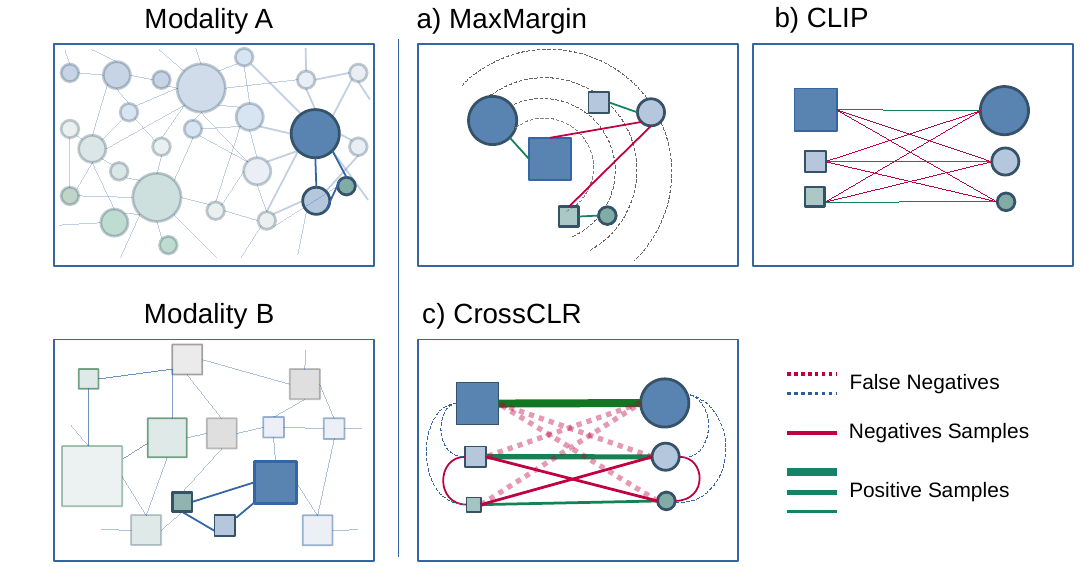}}
    \caption{When learning a joint embedding between two modalities A and B, existing contrastive learning losses, such as MaxMargin~\cite{gabeur2020multimodal,LiuANZ19} and CLIP~\cite{radford2021learning}, ignore the possibility of false negative samples and, thus, push semantically related concepts apart. The proposed CrossCLR loss indentifies influential samples (large circles/boxes), removes them from the negative set and increases their weight in the minibatch. Moreover, CrossCLR adds intra-modal links to the loss.}
    \label{teaser_fig}
    \end{center}
    \vskip -0.4in
\end{figure}
In this paper, we investigate two issues of existing cross-modal contrastive learning techniques. 
\textbf{1.} The cross-modal loss only ensures that the features from the two modalities map to proximate points in the joint embedding, but they lack an explicit measure that also ensures that similar features from the same modality stay close-by in the joint embedding. 
In previous works, it is implicitly presumed that the similarity between modalities, via transitivity, will also preserve similarity within the modality. 
However, it has not been shown that this is the case\footnote{Since many previous works focus on unsupervised feature learning rather than learning joint embeddings, they do not assume that the input embeddings are meaningful, in the sense that semantically related concepts are initially close-by in the feature space. Therefore, preservation of input similarities is meaningless to them. In this work, we do assume that the input embeddings for each modality (e.g. ImageNet pretrained features) already cover some semantics, and we target a joint embedding that leverages these semantics across modalities.}. 
If similar features from the same modality map to far-apart points in the joint embedding, the embedding lacks semantic meaning and, thus, will generalize badly. 
 \textbf{2.} The focus of previous cross-modal contrastive losses is on the definition of positive pairs, whereas negative samples are randomly drawn from the entire distribution. This does not reflect the effect of what we call \emph{influential samples} -- samples that are similar to many other samples and, thus, have a large influence on the shape of the embedding. Marking influential samples as negatives will likely push apart samples that are actually strongly related. 

As a remedy to the first problem, we propose a contrastive loss that enforces the joint embedding to respect the similarity of samples in the original features spaces. 
Moreover, as a remedy to the second problem, we define influential samples as samples with a high connectivity within the dataset and remove these samples from the set of negatives. 
We also introduce a loss weighting based on the connectivity. 
We show that all three measures lead to an improved cross-modal embedding as tested in terms of video-text retrieval and video captioning. 
While this paper focuses on video and text as modalities, we show that the positive effects of the proposed cross-modal loss generalizes to other pairs of modalities.

\section{Related Work}

\subsection{Sample Selection in Contrastive Learning}
Different from the recent research \cite{chen_icml2020_simclr,he_cvpr2020_moco,chen2020improved,chen_arxiv2020_simclrv2,grill_arxiv2020_byol,caron_nips2020_swav}, our work addresses multi-modal contrastive learning. We propose inter- and intra-modality loss objectives to ensure that samples with similar content stay close in the joint embedding space, regardless of the modality.
However, sample selection plays an important role in contrastive learning.
Inspired by Khosla et al.~\cite{khosla2020supervised}, Zhang et al. \cite{zhang_arxiv2020_adaclr} proposed an adaptive self-supervised learning technique to mine nearest positive samples without using label information. 
Han et al. \cite{Han20} introduced a co-training method to mine hard positive samples by using other complementary views of the data for video representation learning.
In terms of negative sampling, recent work explored informative (hard) negatives to facilitate better and faster contrastive learning~\cite{kalantidis_nips2020_hardneg,robinson_iclr2021_hardneg}.

Our work also focuses on selecting better negative samples, but instead of mining hard negatives, we introduce the concept of influential samples. We define these as samples that are strongly connected with others and more likely lead to semantic collision. We exclude them from the negative set and give them more weight in the loss. 

\subsection{Multi-modal Representation Learning}

Video data often consists of multiple modalities, such as raw RGB, motion, audio, text, detected objects, or scene labels.
Employing multiple of these together helps better understand the content of video~\cite{miech19endtoend,gabeur2020multimodal}.
Recently, transformer-based models for cross-modal representation learning became popular~\cite{gabeur2020multimodal,ging2020coot,radford2021learning}.
VideoBERT~\cite{9009570} adopted the BERT design and applied it on quantized video representations.
Miech et al.~\cite{miech19howto100m} introduced the HowTo100M dataset.
Later work exploited the noisy pairings of this dataset for pretraining video-language models~\cite{miech19endtoend,actbert20}. MIL-NCE~\cite{miech19endtoend} proposed to consider multiple positive pairs sampled in a temporally close neighborhood to tackle the misaligned video narrations. 

While the above works focus on learning representations from scratch exploiting very large datasets, another line of research is to learn a joint embedding given pre-trained expert models to compute the input features~\cite{LiuANZ19,gabeur2020multimodal,miech2018learning}. 
Miech et al.~\cite{miech2018learning} computed features from pre-trained experts for text-to-video retrieval. The overall similarity is a weighted sum of each individual expert's similarity score. CE~\cite{LiuANZ19} proposed a mixture of experts model and a collaborative gating mechanism to combine experts.
A recent extension (MMT)~\cite{gabeur2020multimodal} extracts the expert features from 7 different domains and employs a time-aware modality aggregation mechanism. 
MMT incorporates a bidirectional max-margin ranking loss for training.

Our work belongs to the second group of works, where we assume pre-trained input embeddings. \loss overcomes the limitations of max-margin ranking and contrastive learning losses by enforcing consistency within each modality. It avoids pushing away semantically similar representations in the joint embedding space by introducing and exploiting influential samples.

\section{CrossCLR}

In this section, we first define the cross-modal learning task and highlight the issues that normal contrastive learning faces when learning a cross-modal embedding. 
Then we introduce modifications to the contrastive loss to ensure intra-modal alignment and to avoid semantic collision. 

\paragraph{Cross-modal alignment.}
Cross-modal alignment aims to learn two encoders $f_x(\cdot)$ and $f_y(\cdot)$ that map embeddings $x$ and $y$ of two modalities $A$ and $B$
to $z_x = f_x(x)$ and $z_y = f_y(y)$, such that $z_x$ and $z_y$ are close to each other
in the learned embedding space if they refer to similar content, otherwise far away.
That means, it aims to learn a similarity function between samples from $A$ and $B$.
Given a sample pair $(x_i,y_i)$, which is assumed to describe similar content. For example, $x_{i}$ and $y_{i}$ can be the embeddings of a video clip and a corresponding text description of its content, respectively.
A successfully learned cross-modal similarity is supposed to generalize to unseen pairs, notwithstanding the typically large variation in the input modalities $A$ and $B$ even when they show similar content. 
$A$ and $B$ can be arbitrary modalities. 
In this paper, we assume samples from $A$ to be a feature embedding derived from a video clip and samples from $B$ to be a feature embedding of a sentence. 
However, we also show an experimental study where $A$ and $B$ are different modalities derived from a video clip. 
\begin{figure*}[t]
   \vskip 0.2in
   \begin{center}
   \centerline{\includegraphics[width=\textwidth]{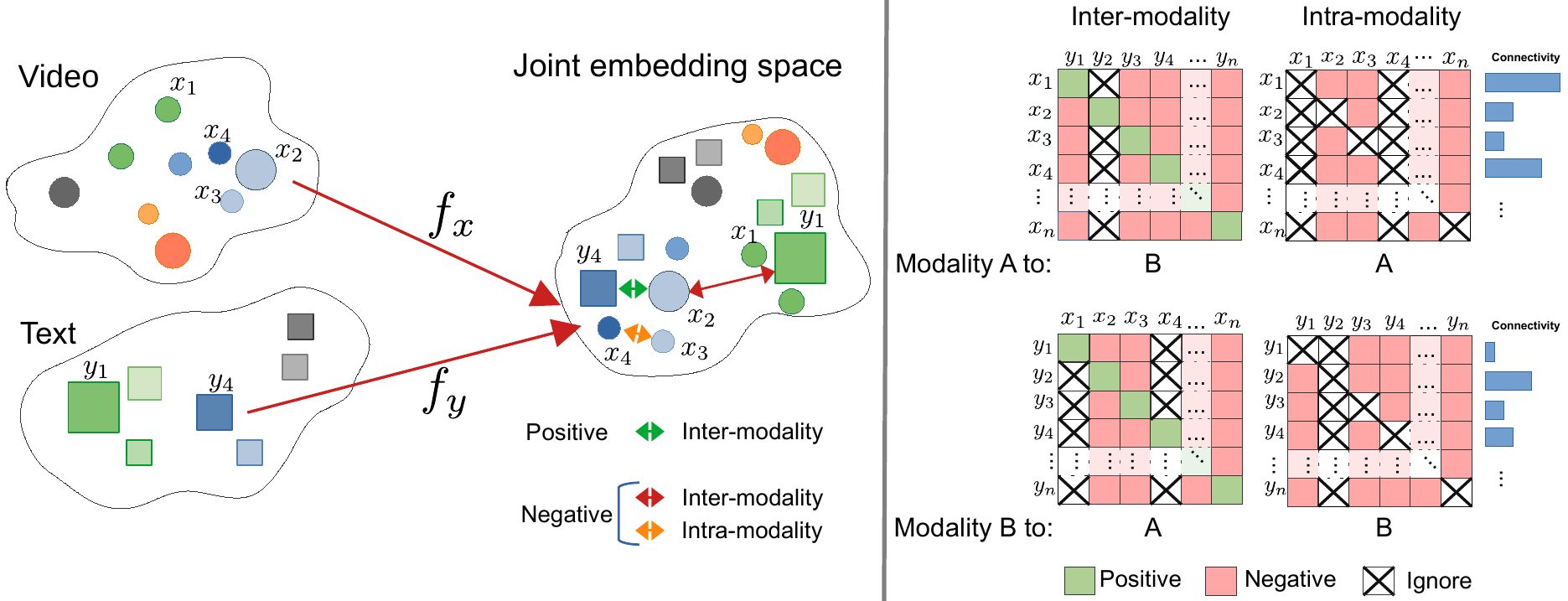}}
   \caption{CrossCLR loss. We first find the influential samples. 
   Influential samples are similar to a large set of other samples. We emphasize these samples in the loss.
    Additionally, we prune these samples from the negative set, because we want to prevent the loss to falsely 
    push them apart while they share semantics with other samples. }
    \label{fig:cross_clr_fig}
   \end{center}
   
   \vskip -0.4in
\end{figure*}
\paragraph{Pretrained feature experts.} We encode feature representations from videos, such as action, appearance, object using off-the-shelf feature extractors~\cite{gabeur2020multimodal, miech2018learning, mithun2018learning, LiuANZ19}. 
Each expert is a pre-trained model on a specific task, more details in Section~\ref{sec:expert_feat}. Given a video $v$, we sample $m$ frames and feed them to the expert model to obtain per-frame expert features $x=[e_1,\dots,e_m]$.
For text data, we use \emph{BERT-Base uncased} model as feature extractor. 

To align embeddings of modality $A$ to embeddings of modality $B$, 
we follow the two-stream hierarchical architecture of COOT~\cite{ging2020coot}, as shown in Fig.\ref{net_fig}. 
It consists of a local transformer for clip-level embeddings and a global transformer for video-level embeddings. 
Given frame/word-level expert features, we obtain the clip/sentence level embeddings via the local 
transformer. These local embeddings serve as input to the global transformer, which yields the final video/paragraph representation. 
\vspace{-0.1in}
\paragraph{Contrastive learning.}
The sample pairs $(x_i,y_i)$ give us information for contrastive learning. 
In particular, each such sample $i$ can be regarded as a positive pair, whereas all pairs $(x_i,y_j)$ for $j\neq i$ will be regarded as negative pairs. To be precise, for each sample $x_i$ in 
the minibatch $\mathcal M$, the positive set $\mathsf{P}_i$ and negative set $\mathsf{N}_i$ are defined as $\mathsf{P}_i = \{y_i\}$ and $\mathsf{N}_i = \{y_j |\forall y_j \in \mathcal M, j\neq i\}$.
The corresponding contrastive loss on a minibatch $\mathcal M$ is: 
\begin{equation}
   \resizebox{.9\hsize}{!}{ $
   L =     \mathbb{E}_{i\in\mathcal M} \left [-\log \frac{\exp(f_x(x_i)^T f_y(y_i))}{\exp(f_x(x_i)^T f_y(y_i)) + {\sum\limits_{y_j\in\mathsf N_i} \exp(f_x(x_i)^T f_y(y_j))}} \right ],
       $}    \label{eq_ideal}
\end{equation}
which is being optimized w.r.t. the parameters of the encoders $f_x$ and $f_y$. 

One weakness in this formulation is the ad-hoc definition of negative samples. 
We simply assumed that all combinations $(x_i, y_j)$ with $i\neq j$ have dissimilar content. 
However, this is not always the case. 
For instance, two sentences ``someone drinks a cup of tea'', and ``she talks on the phone while drinking tea'', aligned with two different video clips $i$ and $j$ have largely overlapping content. 
The loss objective in equation (\ref{eq_ideal}) will only be minimized if $f_x(x_i)$ and $f_y(y_j)$ are mapped to different points in the embedding, despite their high semantic similarity. 
Moreover, nothing in the above loss enforces that $f_y(y_i)$ and $f_y(y_j)$ are mapped to proximal points in the joint embedding space, even when $y_i$ and $y_j$ are similar in their original embedding space, as in the given example.  

\vspace{-0.1in}
\subsection{Inter-Modality and Intra-Modality Alignment}

Let us first approach the missing alignment of the input modalities in the joint embedding. 
If we assume, like in this paper, that the input modalities have already passed an encoder network that is able to place semantically related inputs nearby in the original feature embedding, 
we should preserve this proximity also in the joint embedding space. 
Therefore, $f_x(\cdot)$ and $f_y(\cdot)$ should be optimized not only to map $x_i$ and $y_i$ of a pair $(x_i,y_i)$ to a proximate location in the joint embedding (inter-modality),
 but similar samples $x_i$ and $x_j$ from the same modality should be mapped in close proximity (intra-modality).
 The inter-modality and intra-modality negative sets for sample $x_i$ are defined as: 
 $\mathsf{N}^{E}_i = \{y_j |\forall y_j \in \mathcal M, j\neq i\}$ and $\mathsf{N}^{R}_i = \{x_j |\forall x_j \in \mathcal M, j\neq i\}$.

Therefore, the learning objective is based on four contrastive components including $A$-to-$B$, $A$-to-$A$, $B$-to-$A$, and $B$-to-$B$ alignments, 
as shown in Fig.~\ref{fig:cross_clr_fig}-Right:
\begin{equation}
   \resizebox{.9\hsize}{!}{ 
  $L(x_i) = - \log \frac {\delta(x_i, y_i)} {\delta(x_i, y_i) + \underbrace{\sum_{y_k \in \mathsf{N}^{E}_i}
  \delta(x_i, y_j)}_{\text{Inter modality negative pairs}} +\lambda \underbrace{\sum_{x_j \in \mathsf{N}^{R}_i} \delta(x_i, x_j)}_{\text{Intra modality negative pairs}}}, $}
\label{eq_fx}
\end{equation} 
\begin{equation}
   \resizebox{.9\hsize}{!}{ 
  $L(y_i) = - \log \frac {\delta(y_i, x_i)} {\delta(y_i, x_i) + \underbrace{\sum_{x_k \in \mathsf{N}^{E}_i}
  \delta(y_i, x_j)}_{\text{Inter modality negative pairs}} +\lambda \underbrace{\sum_{y_j \in \mathsf{N}^{R}_i} \delta(y_i, y_j)}_{\text{Intra modality negative pairs}}}, $}
\label{eq_fy}
\end{equation} 
where $\delta(x_i, y_j) = \exp({f_x(x_i)^T f_y(y_j)/\tau}) = \exp({z_{x^i}^T z_{y^j}/\tau})$.
The second and the third term in the denominator sum up inter-modal and intra-modal negative samples, respectively. $\lambda$ is a hyper-parameter to control intra-modality alignment.
We apply $\ell_2$-normalisation to the input feature embeddings before computing the inner product \cite{8953619,Xinshao19,radford2021learning}.
In such case, the inner product is equivalent to cosine similarity. 
While the nominator is symmetric, the denominator is not. 
Hence, we add up two losses for each sample pair $(x_i,y_i)$ -- one for each modality.

\subsection{Avoiding Semantic Collision}
\label{sec:semantic_collision}
The second issue we identified with regular contrastive learning is the contrasting of false negative samples that have actually strong semantic overlap.
The common assumption in contrastive learning is that a large enough number of negative samples help to learn better representations, because in each training batch, the model contrasts more semantic representatives.
However, Arora et al.~\cite{arora2019theoretical} showed that this assumption does not always hold. 
\begin{figure}[t]
   \begin{center}
   \centerline{\includegraphics[width=\columnwidth]{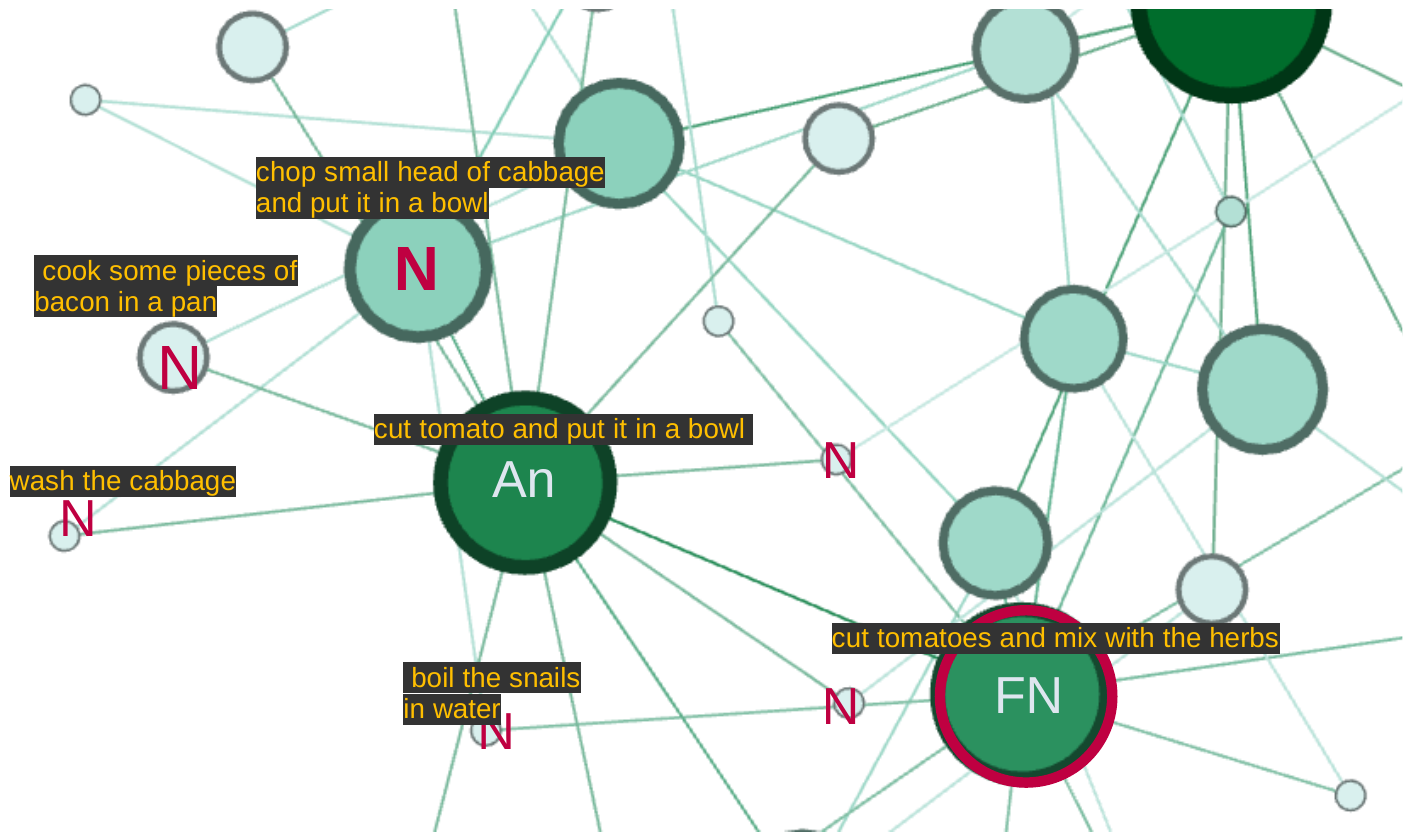}}
   \caption{Example of influential text samples in the Youcook2 dataset (darker green -> more vital). An: Anchor sample, FN: False negative, N: Negative sample. } 
   \label{fig:is_fig}   
\end{center}
   
   \vskip -0.2in
\end{figure}
When there is a large number of negative samples, the chance of observing negative samples with high semantic overlap increases. As shown in Figure~\ref{fig:is_fig}, both samples
\textit{"cut tomato and put it in a bowl"} and \textit{"cut tomatoes and mix with the herbs"} are considered negative samples in standard contrastive learning methods.
By contrasting these undesirable negative pairs, the network is encouraged to discard their common features
in the learned embedding, which is, in fact, the common semantic content, e.g., similar objects and actions in two videos; see also Figure~\ref{fig:is_fig}.
We call this issue \textit{semantic collision}, which is known as ``class collision'' 
in Arora et al. \cite{arora2019theoretical, sohn2020learning} and ``sampling bias'' in Chuang et al. \cite{chuang2020debiased} in a different context. 
When there is a large number of negative samples, frequent semantic collisions can prevent the algorithm from learning good representations.

We argue that it is important to reduce the effect of semantic collision and remove false negatives from the negative set. 
This is non-trivial, since we do not have direct information, e.g., which of the samples are false negatives. 
To this end, we introduce the concept of \textit{influential samples} and propose two components based on influential samples: negative set pruning and loss weighting.

\begin{algorithm}[!t]
    \caption{\label{alg:main} CrossCLR's learning algorithm.}
   
    \begin{algorithmic}
        \STATE \textbf{input:} batch size $N$, queue $Q$, constants $\tau$, $\kappa$, and $\lambda$, networks $f_x$ and $f_y$ for modality A and modality B.
        \STATE \textbf{define} $\delta(x_i, y_j)$ \textbf{as:}
        \STATE $~~~$$\delta(x_i, y_j) = e^{(\frac{{f_x(x_i)^T f_y(y_j)}}{\tau})} = e{(\frac{z_{x_i}^T z_{y_j}}{\tau})}$ 
        \FOR{sampled minibatch $\{\bm x_i,\bm y_i\}_{i=1}^N$}
        \STATE Queuing $\{\bm x_i,\bm y_i\}_{i=1}^N$ in $Q$ and dequeuing oldest keys;
        \STATE \textbf{for all} $k\in \{1, \ldots, |Q|\}$ \textbf{do}
            \STATE $~~~$$c_{x_k} =  \frac{1}{|Q|}\sum\limits_{x_j \in Q}{\frac{x_k^\top x_j}{||x_k||\cdot||x_j||}}$
            \STATE $~~~$$c_{y_k} =  \frac{1}{|Q|}\sum\limits_{y_j \in Q}{\frac{y_k^\top y_j}{||y_k||\cdot||y_j||}}$ 
            \STATE $~~~$ \textbf{if} $c_{x_k} < \gamma$ \textbf{then:} $\hat{\mathsf{N}}^{IR}_{x} \leftarrow\hat{\mathsf{N}}^{IR}_{x}\cup x_k$ \textbf{end if}
            \STATE $~~~$ \textbf{if} $c_{y_k} < \gamma$ \textbf{then:} $\hat{\mathsf{N}}^{IR}_{y} \leftarrow\hat{\mathsf{N}}^{IR}_{y}\cup y_k$ \textbf{end if}
        \STATE \textbf{end for}
        \STATE \textbf{for all} $i\in \{1, \ldots, N\}$ \textbf{do}
            \STATE $~~~$$\alpha _{x_i} =  \frac{1}{|Q|}\sum\limits_{x_j \in Q}{\frac{x_i^\top x_j}{||x_i||\cdot||x_j||}}$
            \STATE $~~~$$\alpha_{y_i} =  \frac{1}{|Q|}\sum\limits_{y_j \in Q}{\frac{y_i^\top y_j}{||y_i||\cdot||y_j||}}$

            \STATE  $~~~$\textbf{if} $\alpha_{x_i} < \gamma$ \textbf{then:} $\hat{\mathsf{N}}^{IE}_{x} \leftarrow\hat{\mathsf{N}}^{IE}_{x}\cup y_i$ \textbf{end if}
            \STATE  $~~~$\textbf{if} $\alpha_{y_i} < \gamma$ \textbf{then:} $\hat{\mathsf{N}}^{IE}_{y} \leftarrow\hat{\mathsf{N}}^{IE}_{y}\cup x_i$ \textbf{end if}

            \small{
            \STATE $~~~$$w_{x_i} = e^{(\frac{\alpha_{x_i}}{\kappa})}; w_{y_i} = e^{(\frac{\alpha_{y_i}}{\kappa})}$
            \STATE $~~~$$L_{x_i} = -  w_{x_i}\log \frac {\delta(x_i, y_i)} {\delta(x_i, y_i) + \sum\limits_{y_k \in \hat{\mathsf{N}}^{IE}_{x}} 
  \delta(x_i, y_k) +\lambda \sum\limits_{x_k \in \hat{\mathsf{N}}^{IR}_{x}} \delta(x_i, x_k)}  $}
            \small{
            \STATE $~~~$$L_{y_i} = -  w_{y_i}\log \frac {\delta(y_i, x_i)} {\delta(y_i, x_i) + \sum\limits_{x_k \in \hat{\mathsf{N}}^{IE}_{y}} 
            \delta(y_i, x_k) +\lambda \sum\limits_{y_k \in \hat{\mathsf{N}}^{IR}_{y}} \delta(y_i, y_k)}  $}

        \STATE \textbf{end for} 
        \STATE update networks $f_x$ and $f_y$ to minimize $L=\frac{L_x+Ly}{2}$
        \ENDFOR
        \STATE \textbf{return} encoders $f_x(\cdot)$ and $f_y(\cdot)$
    \end{algorithmic}
    
    \end{algorithm}

\paragraph{Influential samples.} 
We assume that samples that are strongly connected with other samples are more likely to share semantics and, thus, more likely lead to semantic collision.  
We use a queuing technique to store data samples. This allows us to compute more reliable semantic similarity scores. 
Also recent studies~\cite{8578491,chen2020improved,he_cvpr2020_moco}, showed that a large set of negatives is critical in
contrastive representation learning. The queue size can be much larger than the mini-batch size and we update it progressively by replacing the oldest mini-batch with the current mini-batch.
Hence, given a collection of $M$ samples in queue $\mathcal{Q}_x = \{ x_n\}_{n=1}^M$, 
we define an influential sample as a sample $x_i$ that is strongly connected with many other samples in $\mathcal{Q}_x$, 
where we measure connectivity $C(x_i)$ by the aggregated similarity of features:
\begin{equation}
   C(x_i) = \frac{1}{M}\sum_{j=1}^M{\frac{x_i^\top x_j}{||x_i||\cdot||x_j||}}
\end{equation}

The higher the connectivity, the more influential the data sample is.
Influential samples tend to be either in the center of a semantic cluster or establish a link between clusters, as shown in Fig.~\ref{fig:is_fig}.
We use the connectivity of each sample for pruning and weighting.

\paragraph{Negative set pruning.} 
The connectivity of samples can be thresholded to identify influential samples $\mathcal{I}_x$. Given a collection of samples $\mathcal{Q}_x$ and threshold $\gamma$, we define $\mathcal{I}_x$ as
 $\mathcal{I}_x = \{x_i | C(x_i) > \gamma, \forall x_i \in \mathcal{Q}_x\}$.
To reduce the described effect of false negatives in contrastive learning, we remove all influential samples (in each modality) from the negative set. 
Therefore, we redefine the inter-modality and intra-modality negative sets as: 
 $\hat{\mathsf{N}}^{E}_{x} = \{y_j |\forall (x_j,y_j) \in \mathcal M, x_j\notin \mathcal{I}_x\}$ and $\hat{\mathsf{N}}^{R}_{x} = \{x_j |\forall x_j \in \mathcal{Q}_x, x_j\notin  \mathcal{I}_x\}$.
This is also illustrated in Fig.\ref{fig:cross_clr_fig}-Right.

\paragraph{Loss weighting.}
Moreover, we suggest using the connectivity to emphasize samples with large aggregated connectivity over those with low connectity. 
Samples with very low connectivity can be regarded as outliers to the dataset. 
They are too sparse to positively influence the shape of the embedding. 
Thus, we reduce their influence on the representation learning. 
At the same time, we increase the weight of influential samples, since the cross-modal information of these samples should have a large impact on the shape of the embedding.   
In particular, for each sample and modality we introduce a weight
\vskip -0.1in
\begin{equation}
  w(x_i) = \exp(C(x_i)/\kappa),
\end{equation}
where $\kappa$ is a hyperparameter. 
While we defined the connectivity, the influential samples, and the weights for modality $A$, the same applies for modality $B$. 

The final CrossCLR loss is $(L_x + L_y)/2$ with
\begin{equation}
   \resizebox{.9\hsize}{!}{ 
  $L_x = - \mathbb{E}_{i\in\mathcal M} \left [w(x_i)\log \frac {\delta(x_i, y_i)} {\delta(x_i, y_i) + \sum\limits_{y_k \in \hat{\mathsf{N}}^{E}_{x}} 
  \delta(x_i, y_k) +\lambda \sum\limits_{x_k \in \hat{\mathsf{N}}^{R}_{x}} \delta(x_i, x_k)} \right ] $}
\label{eq_fx}
\end{equation} 
\begin{equation}
   \resizebox{.9\hsize}{!}{ 
  $L_y = - \mathbb{E}_{i\in\mathcal M} \left [w(y_i)\log \frac {\delta(y_i, x_i)} {\delta(y_i, x_i) + \sum\limits_{x_k \in \hat{\mathsf{N}}^{E}_{y}}
  \delta(y_i, x_k) +\lambda \sum\limits_{y_k \in \hat{\mathsf{N}}^{R}_{y}} \delta(y_i, y_k)} \right ] $}
\label{eq_fy}
\end{equation}

\section{Experiments}

\subsection{Datasets and Metrics}
We conducted experiments on LSMDC~\cite{7298940} and Youcook2~\cite{ZhXuCoCVPR18} datasets. 

\textbf{LSMDC}~\cite{7298940} contains 118,081 short video clips extracted from 202 movies. Each clip is annotated with a caption, extracted from either
the movie script or the audio description. The test set is composed of 1000 videos, from movies not present in the training set. 

\textbf{Youcook2}~\cite{ZhXuCoCVPR18} contains 2000 videos with a total number of $14k$ clips. This dataset is collected from
YouTube and covers 89 types of recipes. There are $∼9.6k$ clips for training and $∼3.2k$ clips for
validation. For each clip, there is a manually annotated textual description.

\textbf{Evaluation protocol}. We evaluate the learned embeddings on the modality$-$to$-$modality retrieval task in terms
of Recall$@ K$, median rank (MdR), and mean rank (MnR). Given a query, its $K$ nearest neighbours are retrieved from the database. The retrieval is considered successful, if the correct sample is among the $K$ nearest neighbors.
\begin{figure}[t]
\begin{center}
\centerline{\includegraphics[width=\columnwidth]{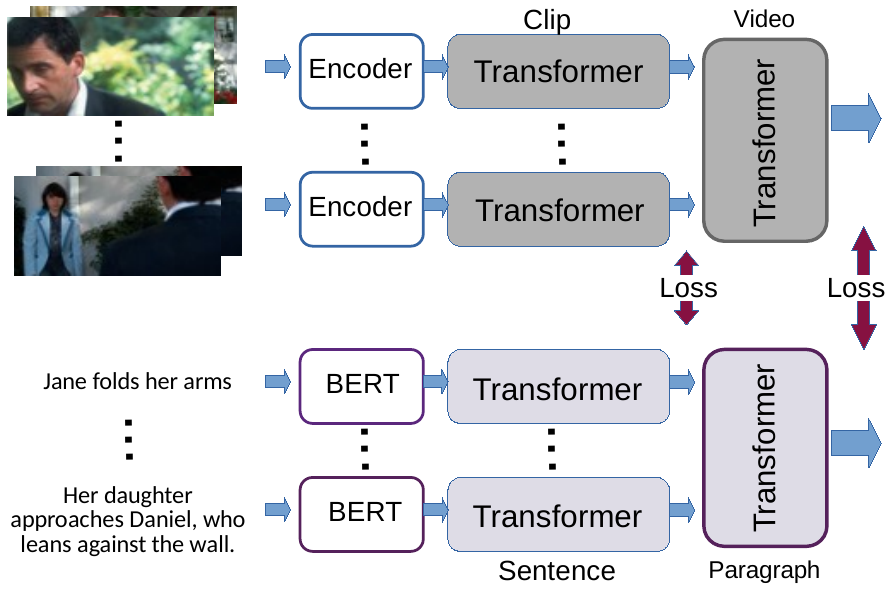}}
\caption{\textbf{Architecture.} The model consist of two branches: one for modality A (e.g. Video) and one for modality B (e.g. Text). Each modality is represented by features from a 
pretrained expert, which we keep frozen (visual Encoder and BERT). These embeddings are fed into a transformer, which maps the input features into a joint embedding space. 
For video and text, we use a two-level hierarchy of transformers, where the loss is applied at the clip/sentence-level and at the video/paragraph level. 
The second stage takes the features from the joint embedding of the first transformer as input.
}
\label{net_fig}
\end{center}
\vskip -0.3in
\end{figure}

\begin{table*}[t]
    \caption{\textbf{Comparison among contrastive learning losses.} Video-text retrieval results with different contrastive learning losses on the YouCook2 and LSMDC dataset. \loss shows consistently higher retrieval scores than previous losses.} 
    \resizebox{2.1\columnwidth}{!}{
    \begin{tabular}{lcccccc||cccccc}
        & \multicolumn{6}{c}{Youcook2} & \multicolumn{6}{c}{LSMDC} \\
        \cmidrule(rl){2-7}  \cmidrule(rl){8-12} 
        & \multicolumn{3}{c}{Text$\implies$Video} & \multicolumn{3}{c}{Video$\implies$Text} & \multicolumn{3}{c}{Text$\implies$Video} & \multicolumn{3}{c}{Video$\implies$Text} \\
         \cmidrule(rl){2-4}  \cmidrule(rl){5-7} \cmidrule(rl){8-10}  \cmidrule(rl){10-12}
          & R@1 & R@5 & R@10  & R@1 & R@5 & R@10 & R@1 & R@5 & R@10  & R@1 & R@5 & R@10  \\ \thickhline
            MaxMargin~\cite{gabeur2020multimodal,LiuANZ19}   & 15.0\std{0.34} & 37.0\std{0.38} & 49.1\std{0.51} & 14.2\std{0.14} & 35.3\std{0.02} & 47.2\std{0.08}               & 8.2 & 22.7 & 31.8 & 9.1 & 23.3 & 31.8\\
            MIL-NCE~\cite{miech19endtoend} &  18.0\std{0.23} & 41.9\std{0.53} & 53.9\std{1.06} & 16.4\std{0.51} & 41.5\std{0.24} & 54.1\std{0.54}         & 8.9 & 24.2 & 32.5 & 10.4 & 25.4 & 34.9 \\
            CLIP~\cite{radford2021learning} &  17.8\std{0.40} & 42.1\std{0.78} & 54.4\std{0.66} & 17.0\std{0.54} & 42.0\std{0.48} & 55.0\std{0.58}        &  9.7 & 24.1 & 32.6 & 9.5 & 23.8 & 32.5 \\
            DCL~\cite{chuang2020debiased} &  17.9\std{0.88} & 41.5\std{0.59} & 54.1\std{0.79} & 16.8\std{0.39} & 42.0\std{0.37} & 55.3\std{0.62}          & 9.0 & 24.9 & 33.2 & 8.6 & 23.4 & 32.2 \\
            NT-Xent~\cite{chen2020simple} &  17.5\std{0.44} & 42.4\std{0.27} & 55.0\std{0.77} & 17.3\std{0.58} & 41.6\std{0.89} & 54.6\std{1.0}            & 9.3 & 23.6 & 32.5 & 10.0 & 25.1 & 33.4  \\
            \hline
            \loss  & \textbf{19.5\std{0.49}} & \textbf{45.9\std{0.55}} & \textbf{58.3\std{0.76}} & \textbf{18.5\std{0.32}} & \textbf{44.8\std{0.82}} & \textbf{57.9\std{0.77}}      & \textbf{10.9} & \textbf{26.2} & \textbf{34.7} & \textbf{12.0} & \textbf{26.1} & \textbf{35.3}\\
    \end{tabular}
    }
    \label{tab:youcook2_loss}
\end{table*}






\subsection{Expert Features}
\label{sec:expert_feat}
We encode the content of a video with pre-trained models trained for different semantic tasks, namely appearance, scene,
action, object, and howto100m~\cite{miech19endtoend}. We extract per-frame features.
To be specific, we use a ResNeSt269 model~\cite{zhang2020resnest} pretrained on ImageNet to extract appearance information,
and a DenseNet161 model~\cite{zhou2017places} pretrained on Places365 to extract scene information.
In terms of action information, we adopt a R(2+1)D model~\cite{tran2019video} with ResNet152 backbone pretrained on IG65M, and
extract the final global pooled feature. For object features, we use a Faster-RCNN model~\cite{ren2016faster} with ResNet50 FPN backbone.
 
For Youcook2 experiments we use the Howto100m features provided by \cite{ging2020coot}. For LSMDC results in the Table~\ref{tab:youcook2_loss}, we used action and appearance features as input to the model. For the SOTA comparisons in Table~\ref{tab:lsmdc_sota} we utilized all expert features including appearance, action, scene, object and howto100m features.
\begin{table}[t]
    \caption{\textbf{Ablation study on CrossCLR loss.} We quantify the individual contributions of the
    \loss components: proximity weighting ($P_W$), intra-modality alignment ($I_M$), and negative pruning ($N_P$) (reported avg and std over 5 runs). }
    \resizebox{1\columnwidth}{!}{
    \begin{tabular}{lccccccccc}
         \multirow{2}{*}{$P_W$} & \multirow{2}{*}{$I_M$} & \multirow{2}{*}{$N_P$} &
        \multicolumn{3}{c}{Text$\implies$Video} & \multicolumn{3}{c}{Video$\implies$Text} 
        \\
         \cmidrule(rl){4-6}  \cmidrule(rl){7-9}
         & & & R@1 & R@5 & R@10  & R@1 & R@5 & R@10   \\ \hline
       
         \xmark & \xmark & \xmark   & 17.9\std{0.51} & 41.7\std{0.83} & 53.7\std{0.77} & 16.7\std{0.53} & 40.9\std{0.64} & 53.7\std{0.96}  \\
        \cmark & \xmark & \xmark   & 18.2\std{0.39} & 41.9\std{1.28} & 53.7\std{1.12} & 16.9\std{1.18} & 41.4\std{1.04} & 53.9\std{1.33}  \\
         \xmark & \cmark & \xmark     &18.7\std{0.33} & 43.6\std{0.63} & 56.2\std{0.54} & 17.7\std{0.50}  & 42.8\std{0.79} & 56.3\std{0.49}   \\
        \cmark & \cmark & \xmark    & 19.0\std{0.42} & 45.4\std{0.98} & 57.8\std{0.97} & 18.3\std{0.35} & 44.1\std{1.13} & 57.2\std{1.11} \\ \hline
         \cmark & \cmark & \cmark & \textbf{19.5\std{0.49}} & \textbf{45.9\std{0.55}} & \textbf{58.3\std{0.76}} & \textbf{18.5\std{0.32}} & \textbf{44.8\std{0.82}} & \textbf{57.9\std{0.77}} \\
    \end{tabular}
    }
    \label{tab:ablation_loss}
\end{table}
\begin{figure}[b]
    \vskip -0.1in
    \begin{center}
    \centerline{\includegraphics[width=\columnwidth]{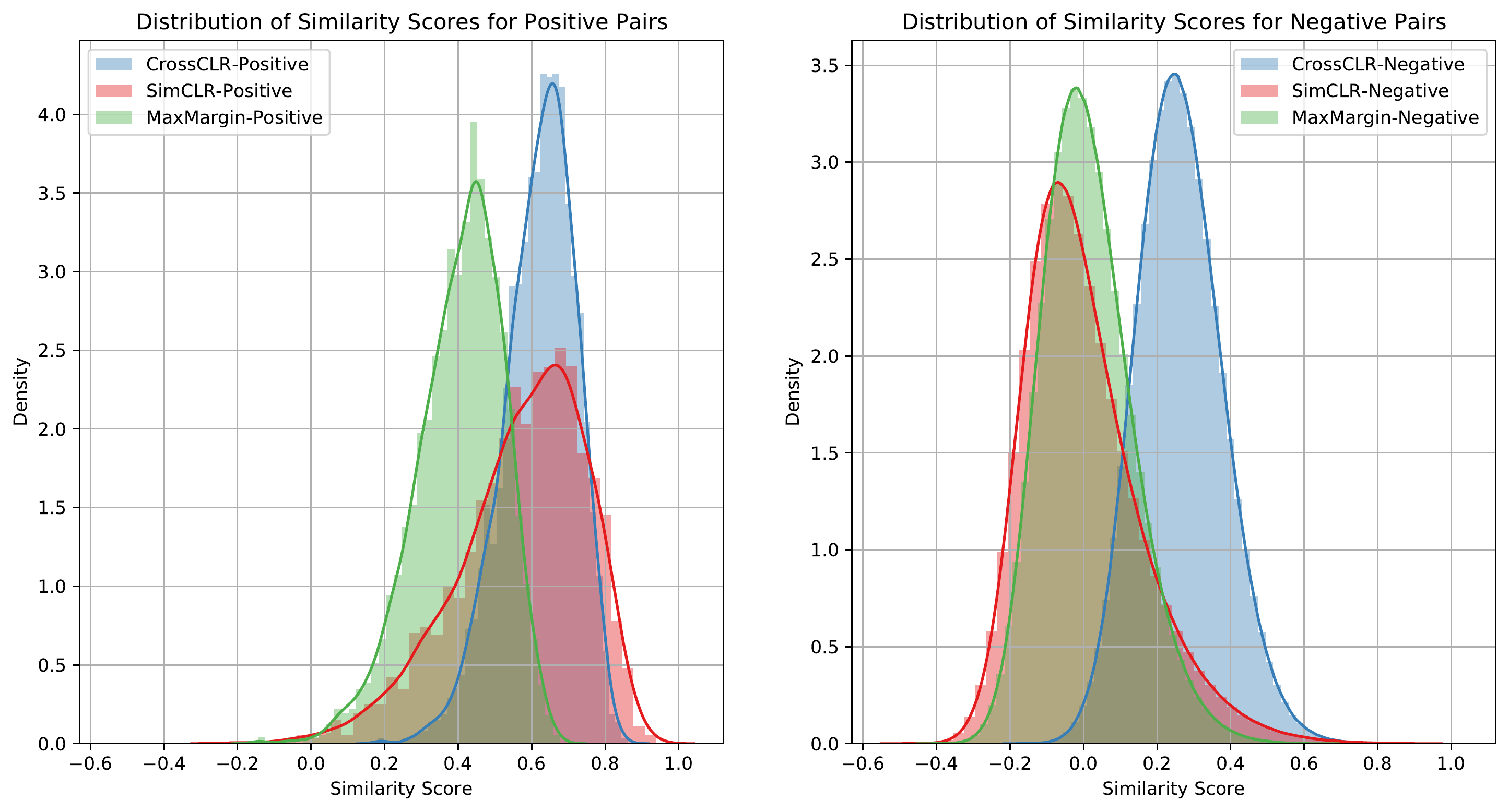}}
    \caption{Distribution of similarity scores for positive pairs and negative pairs in the joint embedding after training for the MaxMargin, 
    NT-Xent (SimCLR), and CrossCLR losses. CrossCLR yields a lower variance (more certainty) on positive pairs compared to MaxMargin and NT-Xent. For MaxMargin, also the mean is shifted towards smaller values, which means it has lower confidence in aligning similar semantics. For negative pairs, NT-Xent and MaxMargin's scores are concentrated around zero. 
     In contrast, CrossCLR intrinsically considers similarity of negative pairs and therefore the concentration is shifted towards positive scores.}
    \label{compare_cosine}
    \end{center}
\end{figure}

\subsection{Implementation Details}

We use the RADAM optimizer with momentum 0.56 and mini-batch size of 64. The initial learning rate is 
set to $7e4$. It is reduced by a factor of 10 when validation performance saturates. For each dataset, we use the corresponding validation set to estimate the hyperparameters.
We use the following hyperparameters for the \loss: $\kappa = 35e4$ for Youcook2 and $\kappa = 55e4$ for LSMDC , $\lambda = 8e1$ for Youcook2 and $\lambda = 65e2$ for LSMDC, $\tau=0.03$. 
We use threshold $\gamma=0.9$ and queue size of $3K$ and $5K$ for Youcook2 and LSMDC datasets respectively.
We train the models for 40 epochs which take $\sim$1 hour for Youcook2 and $\sim$2 hour for LSMDC datasets to train on one NVIDIA T4 Tensor core GPU.
For further details on the architecture, we refer to the supplementary material.

\subsection{Experiment Results}

\begin{figure*}[h]
    \begin{center}
    \centerline{\includegraphics[width=2.1\columnwidth]{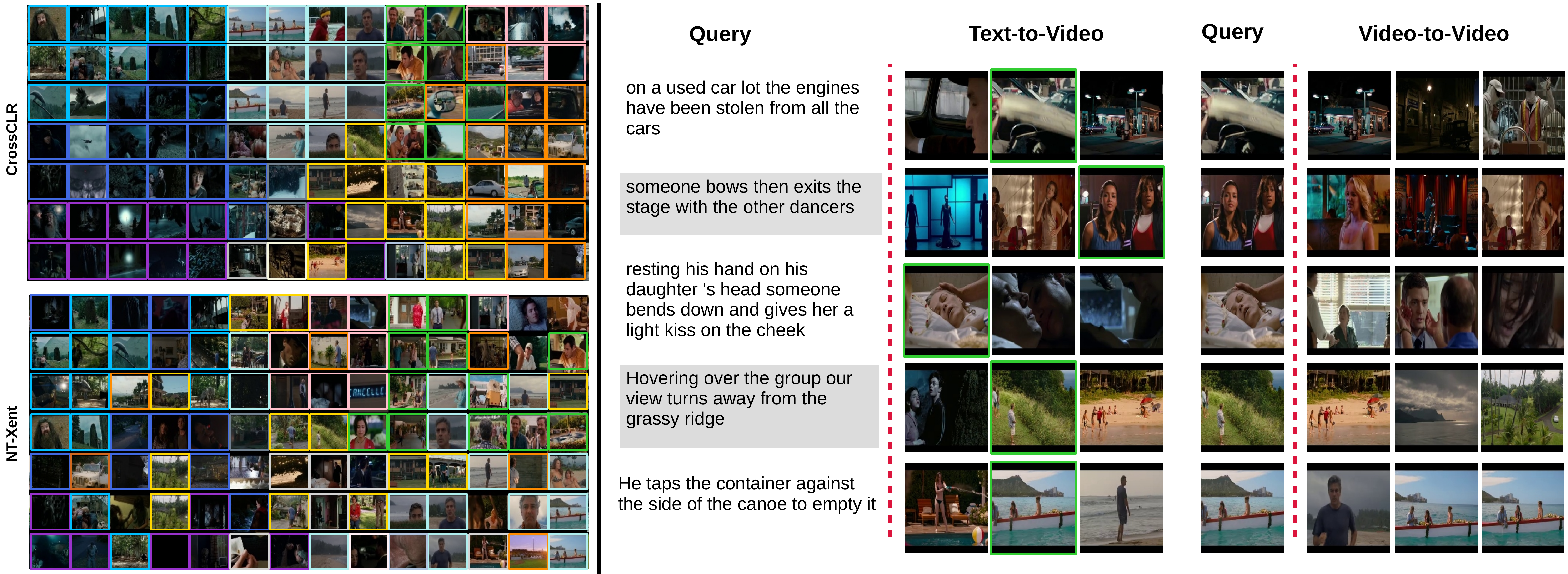}} 
    \caption{Qualitative results for LSMDC dataset (Best viewed in color). \textbf{Left}: We show the t-SNE visualization of embeddings for \loss and NT-Xent based on text embeddings.
    Color of the boxes represent the similarity of semantics. CrossCLR has higher intra-modality consistency than NT-Xent. \textbf{Right}: It shows several examples 
    for text-to-video (Top3 nearest neighbors) and video-to-video (Top3 nearest neighbors excluding self) retrieval task.}
    \label{retrieval_qual}
    \end{center}
    \vskip -0.3in
\end{figure*}

\textbf{Ablation study of CrossCLR components.}
The proposed \loss contains three components: the intra-modality alignment ($I_M$), proximity weighting ($P_W$), and negative pruning ($N_P$). 
To shed light on the contributions of these three components, we report the retrieval results on the Youcook2 dataset in Table~\ref{tab:ablation_loss}.
The data split and other experimental settings were kept identical.
All three components have a positive impact on the results, individually and jointly. 
The intra-modality component $I_M$ yields the largest individual gain, but also the concept of influential samples, as used by negative pruning and proximity weighting, yields a clear benefit. 

\textbf{Comparison to other contrastive learning methods.}
We compare the \loss loss with popular contrastive losses: MaxMargin~\cite{triplet1, triplet2,gabeur2020multimodal,LiuANZ19}, MILNCE~\cite{miech19endtoend}, 
NT-Xent~\cite{chen_icml2020_simclr}, CLIP~\cite{radford2021learning}, and DCL~\cite{chuang2020debiased}. 
For the sake of a fair and direct comparison, we used the same experimental setup and architecture as COOT~\cite{ging2020coot} and only exchanged the loss. 
Table~\ref{tab:youcook2_loss} reports the retrieval performance on Youcook2 and LSMDC.
\loss consistently improves over previous contrastive learning methods on both datasets.
\begin{table}[t]
    \caption{\textbf{Comparison to the state-of-the-art text-to-video retrieval on Youcook2.} \loss improves the state of the art (COOT) while using the same architecture and experimental settings.}
    \small
    \resizebox{1\columnwidth}{!}{
    \begin{tabular}{lccc}
        & \multicolumn{3}{c}{Text$\implies$Video} \\
         \cmidrule(rl){2-4}  
          & R@1 & R@5 & R@10   \\ \thickhline
            Random             & 0.03      & 0.15              & 0.3   \\ 
            Miech et al.~\cit{miech19howto100m} & 6.1               & 17.3              & 24.8  \\
            ActBERT~\cit{actbert20} & 9.6               & 26.7              & 38.0   \\
            MIL-NCE~\cit{miech19endtoend} & 15.1              & 38.0              & 51.2   \\
            COOT~\cite{ging2020coot}  & 16.7\std{0.4}   & 40.2\std{0.3}   & 52.3\std{0.5}   \\ 
            \hline
            \loss  & \textbf{19.5\std{0.49}} & \textbf{45.9\std{0.55}} & \textbf{58.3\std{0.76}}  \\
    \end{tabular}
    }
    \label{tab:youcook2_sota}
    \vskip -0.1in
\end{table}

\begin{table}[t]
    \caption{\textbf{Comparison to the state-of-the-art text-to-video retrieval on LSMDC dataset (test split).} \loss improves the state of the art (MMT).}
    \small
    \resizebox{1\columnwidth}{!}{
    \begin{tabular}{lccccc}
        & \multicolumn{5}{c}{Text$\implies$Video} \\
         \cmidrule(rl){2-6}  
          & R@1 & R@5 & R@10  & MdR$\downarrow$  & MnR$\downarrow$\\ \thickhline
            Random &   0.1 & 0.5 & 1.0 & 500.0 & 500.0  \\ 
            CT-SAN~\cite{yu2017endtoend} &  5.1 & 16.3 & 25.2 & 46.0 & $-$ \\
            JSFusion~\cite{jsf18} &  9.1 &  21.2 &  34.1  & 36.0 & $-$\\
            CCA~\cite{klein2015associating} &  7.5 & 21.7 & 31.0 & 33.0 & $-$ \\
            MEE~\cite{miech2018learning} &  9.3 & 25.1 & 33.4 & 27.0 & $-$ \\
            MEE-COCO~\cite{miech2018learning} &  10.1  & 25.6  & 34.6 &  27 & $-$ \\
            CE~\cite{LiuANZ19} &  11.2  & 26.9  & 34.8 & 25.3 & $-$ \\
            MMT~\cite{gabeur2020multimodal} &  13.2 &  29.2 &  38.8 & 21.0 & 76.3\\ 
            COOT~\cite{ging2020coot}  & 11.3   & 26.1   & 36.7  & 22.0 & 85.4 \\ 
            CLIP~\cite{portilloquintero2021straightforward} &  11.3 &  22.7 &  29.2 & 56.5 & $-$\\ 
            \hline
            \loss  & \textbf{15.0} & \textbf{32.5} & \textbf{42.0} &\textbf{18.0} &\textbf{74.4} \\
    \end{tabular}
    }
    \label{tab:lsmdc_sota}
    \vskip -0.1in
\end{table}

In Figure~\ref{compare_cosine}, we explore different contrastive learning methods,  three contrastive methods SimCLR (NT-Xent), MaxMargin, and \loss,  w.r.t. their similarity scores of positive and negative pairs. 
We show the histograms of similarity scores for positive and negative samples separately. The scores are computed 
on the validation set after training the network with these three losses individually.
Figure~\ref{compare_cosine} reveals that \loss yields a higher certainty (lower variance) for positive samples than the MaxMargin loss or SimCLR loss. The MaxMargin loss~\cite{triplet1, triplet2,gabeur2020multimodal,LiuANZ19} yields a significantly lower mean score for positive samples, which indicates that this method has issues installing the provided training links between cross-modal concepts in the learned embedding. SimCLR~\cite{chen_icml2020_simclr} can represent more of these concepts, but the higher variance indicates that there are many failure cases, too.  
From the histograms over the negative sample scores we can see that SimCLR and MaxMargin loss both have zero mean for negative samples, as enforced by the structure of the loss. This does not allow for semantic overlap among negative samples. 
In contrast, \loss allows for semantic similarity among negative samples, which shifts the distribution towards positive scores.
This does not necessary mean that the relationships between negative samples enabled by \loss are all meaningful. 
However, the improved retrieval performance after pruning of influential samples indirectly indicates that many of these relationships make sense. 
\definecolor{Gray}{gray}{0.92}
\newcolumntype{a}{>{\columncolor{Gray}}c}
\newlength\savewidth\newcommand\shline{\noalign{\global\savewidth\arrayrulewidth
  \global\arrayrulewidth 1pt}\hline\noalign{\global\arrayrulewidth\savewidth}}

\newlength\thinwidth\newcommand\thinline{\noalign{\global\savewidth\arrayrulewidth
  \global\arrayrulewidth 0.5pt}\hline\noalign{\global\arrayrulewidth\savewidth}}

  \definecolor{LightCyan}{rgb}{0.88,0.7,0.5}
\definecolor{highlightRowColor}{gray}{0.82}
\newcommand{\colorrow}{\rowcolor{LightCyan}}
\newcommand{\boldunder}[1]{\textbf{\underline{#1}}}

\begin{table*}[!t]
    \caption{\textbf{Generalization to other pairs of modalities.} Modality-to-modality retrieval on the full grid of modalities for the LSMDC dataset. \loss yields improved retrieval scores for almost all pairs of modalities.}

    \centering
    
    \setlength{\tabcolsep}{0.5em}\resizebox{\linewidth}{!}{
    \begin{tabular}{l|l|aa|cc|aa|cc|aa||cc}
    Modality & Method & \multicolumn{2}{c}{Text} & \multicolumn{2}{c}{Appearance}
    & \multicolumn{2}{c}{Object} & \multicolumn{2}{c}{Scene}  & \multicolumn{2}{c}{Action} & \multicolumn{2}{c}{\textbf{Mean}} \\
     & & {\small{\textbf{R@1}}} & {\small{\textbf{R@10}}} & {\small{\textbf{R@1}}} & {\small{\textbf{R@10}}} & {\small{\textbf{R@1}}} & {\small{\textbf{R@10}}} & {\small{\textbf{R@1}}} & {\small{\textbf{R@10}}}
     & {\small{\textbf{R@1}}} & {\small{\textbf{R@10}}} & {\small{\textbf{R@1}}} & {\small{\textbf{R@10}}}\\
    
    \shline


    \multirow{2}{*}{Text}  & NT-Xent & \cellcolor{gray}$--$ & \cellcolor{gray}$--$ & 7.2 & 28.3 & 2.8 & 8.3 & 5.7 & 24.7 & 9.3 & 32.6 & 6.3 & 23.5\\
    & CrossCLR & \cellcolor{gray}$--$ & \cellcolor{gray}$--$ & 8.1 & 29.3 & 2.5 & 9.6 & 7.4 & 25.3  & 10.2 & 33.3 & \textbf{7.1} & \textbf{24.4} \\
                    
    \thinline

    \multirow{2}{*}{Appearance}  & NT-Xent & 7.1 & 27.5 & \cellcolor{gray}$--$ & \cellcolor{gray}$--$ & 72.7 & 92.2 & 98.3 & 100 & 85.6 & 98.2 & 65.9 & 79.5\\
     & CrossCLR & 8.8 & 30.0 & \cellcolor{gray}$--$ & \cellcolor{gray}$--$ & 75.8 & 93.8 & 99.5 & 100 & 85.4 & 98.9 & \textbf{67.4} & \textbf{80.7}\\
    
    \thinline

    \multirow{2}{*}{Object}  & NT-Xent & 2.4 & 8.3 & 68.6 & 91.5 & \cellcolor{gray}$--$ & \cellcolor{gray}$--$ & 75.4 & 95.2 & 43.2 & 80.5 & 47.4 & 68.9\\
     & CrossCLR & 3.1 & 9.0 & 74.7 & 94.7 & \cellcolor{gray}$--$ & \cellcolor{gray}$--$ & 79.5 & 95.9 & 43.3 & 82.4 & \textbf{50.2} & \textbf{70.5}\\

    \thinline

    \multirow{2}{*}{Scene}  & NT-Xent & 6.2 & 24.3 & 98.5 & 100 & 82.8 & 95.6 & \cellcolor{gray}$--$ & \cellcolor{gray}$--$ & 86.8 & 98.7 & \textbf{68.6} & 79.7\\
     & CrossCLR & 7.4 & 25.0 & 99.6 & 100 & 84.0 & 96.0 & \cellcolor{gray}$--$ & \cellcolor{gray}$--$ & 83.0 & 98.8 & 68.5 & \textbf{80.0}\\

    \thinline

    \multirow{2}{*}{Action}  & NT-Xent & 7.6 & 31.7 & 84.8 & 97.9 & 47.1 & 85.4 & 85.3 & 98.5 & \cellcolor{gray}$--$ & \cellcolor{gray}$--$ & 56.2 & 78.4\\
     & CrossCLR & 9.7 & 33.0 & 85.0 & 98.8 & 50.3 & 86.2 & 86.1 & 98.7 & \cellcolor{gray}$--$ & \cellcolor{gray}$--$ & \textbf{57.8} & \textbf{79.2}\\

    \hline \hline

    \multirow{2}{*}{\textbf{Mean}}  & NT-Xent & 5.8 & 23.0 & 64.8 & 79.4 & 51.4 & 70.4 & 66.2 & 79.6 & \textbf{56.2} & 77.5 & \cellcolor{gray}$--$ & \cellcolor{gray}$--$\\
     & CrossCLR & \textbf{7.3} & \textbf{24.3} & \textbf{66.9} & \textbf{80.7} & \textbf{53.2} & \textbf{71.4} & \textbf{68.1} & \textbf{80.0} & 55.5 & \textbf{78.4} & \cellcolor{gray}$--$ & \cellcolor{gray}$--$\\

    \thinline
    \end{tabular}
    }
    \label{tab:mod_to_mod_lsmdc}
\end{table*}

Figure~\ref{retrieval_qual} shows gualitative visualization of learned embeddings. The qualitative samples in this figure show that \loss brings more consistency over intra-modality samples in comparison to NT-Xent.

\textbf{Comparison to the state of the art.}
Table~\ref{tab:youcook2_sota} and Table~\ref{tab:lsmdc_sota} show that, thanks to the \loss loss, retrieval performance from text to video improves over the state of the art on both the YouCook2 and the LSMDC dataset. For LSMDC experiment we utilized action, appearance, object, scene and howto100m features. However, MMT~\cite{gabeur2020multimodal} utilizes additional modalities such as Audio, OCR and Face.
 \loss outperforms MMT while using less modalities. Note that for LSMDC experiment in Table~\ref{tab:lsmdc_sota}, we don’t use sophisticated fusion techniques and just simply aggregate output of each model trained individually on each modality (more details in supplementary material).
Since, we use the same architecture (Fig. \ref{net_fig}) and experimental setup as COOT~\cite{ging2020coot}, with only the loss being different, the improvement can be directly attributed to \loss. 


\begin{table}[t]
    \caption{
        \tbf{Captioning performance on the YouCook2 dataset.}  We follow the setup from COOT~\cite{ging2020coot} and report captioning results on
the validation split, given ground truth video segments. The last row shows the human performance and can be considered as the best achievable performance as reported by Hessel et al~\cite{hessel_caption19}. $^+$ We use video embeddings in addition to clip embeddings as input to model. $^\star$ Vedantam et al.~\cite{cider} introduced two versions CIDEr and CIDEr-D. We report CIDEr-D~\cite{cider}, however for methods with $^\star$ sign we couldn't find the information about the version they used. }
    \centering
    \resizebox{1\columnwidth}{!}{
    \begin{tabular}{lll llll llll}
        \hline
Method                  & B$@$3    & B$@$4     & RougeL   & METEOR         & CIDEr-D      & RE$@$4$\downarrow$\\\hline
VideoBERT~\cite{sun2019videobert}                  &  7.59 &         4.33 &     28.80 &     11.94 &     $-^\star$ & $-$\\
ActBERT~\cite{actbert20}                  &     8.66 &      5.41 &     30.56 &     13.30 &     $-^\star$ & $-$\\
Zhou et al.~\cite{zhou2018endtoend}         &  7.53         &           3.84 &     27.44 &     11.55 &     $-^\star$ & $-$\\
VTransformer~\cite{zhou2018endtoend}      &    13.08&     7.62 &     32.18&     15.65 &     32.26 &      7.83 \\
TransformerXL~\cite{dai2019transformerxl}       &   11.46&       6.56 &     30.78&     14.76 &     26.35 &      6.30 \\
MART~\cite{mart}                  &      12.83&     8.00 &     31.97&     15.90 &     35.74 & 4.39\\
AT+Video~\cite{hessel_caption19} &   $-$       &           9.01 &     36.65 &     17.77 &     $-^\star$ & $-$\\

\hline

COOT~\cite{ging2020coot}                    &    17.12 &     10.91 &     37.59 &     18.85 &     54.07 &      5.11 \\
CrossCLR        & 17.60  &    11.11 &     38.00 &     19.25 &     58.65 &      \textbf{4.19} \\
\hline

COOT~\cite{ging2020coot}$^+$     & 17.97        & 11.30 & 37.94 & 19.85 & 57.24 &      6.69 \\
CrossCLR$^+$           & \textbf{18.62}  &    \textbf{12.04} &     \textbf{38.63} &     \textbf{20.17} &     \textbf{61.10} &     5.62 \\

\hline
Human~\cite{hessel_caption19}            & $-$  &    15.20 &     45.10 &     25.90 &     $-$ &     $-$ \\
 
\hline
    \end{tabular} } 
    \label{tab:captioning_yc}
\end{table}

\textbf{Other pairs of modalities.}
We were interested whether the improvements obtained by \loss are specific to video-text retrieval or if the loss generalizes well to other pairs of modalities. 
To this end, we compared \loss to the NT-Xent loss~\cite{chen2020simple} on various combinations of input modalities derived from the LSMDC dataset. 
Table~\ref{tab:mod_to_mod_lsmdc} shows that, with very few exceptions, \loss shows better retrieval performance than NT-Xent. This demonstrates that the principles behind \loss are not specific to video-text retrieval but are applicable to learning cross-modal embeddings in general. 

\subsection{Video Captioning}
In addition to the video-text retrieval, we report the captioning performance to show the richness of learned embeddings with CrossCLR. Table~\ref{tab:captioning_yc} reports the captioning performance of learned embeddings with CrossCLR method in comparison to state-of-the-art methods on Youcook2 validation set. 

We evaluate the captioning performance similar to MART~\cite{mart} and COOT~\cite{ging2020coot} with the metrics CIDEr-D~\cite{cider}, ROUGE-L~\cite{lin-2004-rouge}, BLEU@3 and BLEU@4~\cite{papineni-etal-2002-bleu}, and METEOR~\cite{denkowski-lavie-2014-meteor}. Additionaly, we utilize repetition metric RE@4~\cite{ngramrepetition} to measure the reptition of the caption. We feed the clip embeddings generated with our CrossCLR method to the captioning model MART~\cite{mart} (additionally we also use video embeddings for $^+$ experiment). In comparison to COOT we obtain more diverse and accurate captions. MART~\cite{mart} uses appearance and optical flow features extracted from ResNet-200 and BN-Inception models, respectively.
According to all metrics, CrossCLR outperforms state-of-the-art significantly and also generates more diverse captions compared to other methods.


\section{Conclusion}

We introduced a contrastive loss for learning joint embeddings of two input modalities that respects the special need of cross-modal learning. First, assuming that the input modalities already come with some meaningful embedding where semantically related concepts are close in feature space, we enforced this proximity also in the joint embedding. Previous losses did not use this constraint, partially because they primarily focus on learning feature representations of the input modalities from scratch rather than learning a joint embedding. In our experiments, the intra-modal proximity led to consistent benefits in cross-modal retrieval performance. Second, we identified the problem of false negative samples and proposed a heuristic to identify false negatives via the degree of interconnections in the input embedding. Also this, together with the proximity weighting, led to consistent improvements in retrieval performance. Finally, we showed that the improvements obtained with \loss are not limited to video and text embeddings. The same principle can be successfully applied to other pairs of modalities to learn joint embeddings.  

\begin{appendices}
\section{Architecture}

A video $v_i$ is represented as a collection of consecutive clips $vi = [c_{i1},c_{i2},\dots,c_{il}]$, and similarly we define paragraph $p_i$ as a list of sentences $p_i=[s_{i1},s_{i2},\dots, s_{il}]$. Each pair video $v_i$ and paragraph $p_i$ and also their clips ($c_{ij}$) and sentences ($s_{ij}$) are considered temporally aligned.

We first apply pre-trained visual encoders (appearance, object, etc) to extract per-frame features, where a clip $c_{ij}$ contains $t$ frames. For the sentence $s_{ij}$ with $h$ words, we utilize \textit{Bert-Based uncased} model to extract features.
As explained in Section 4.3, for Youcook2 we use HowTo100m pre-trained model~\cite{ging2020coot} to encode video frames.

\begin{figure}[t]
    \centering
    \includegraphics[width=1.0\columnwidth]{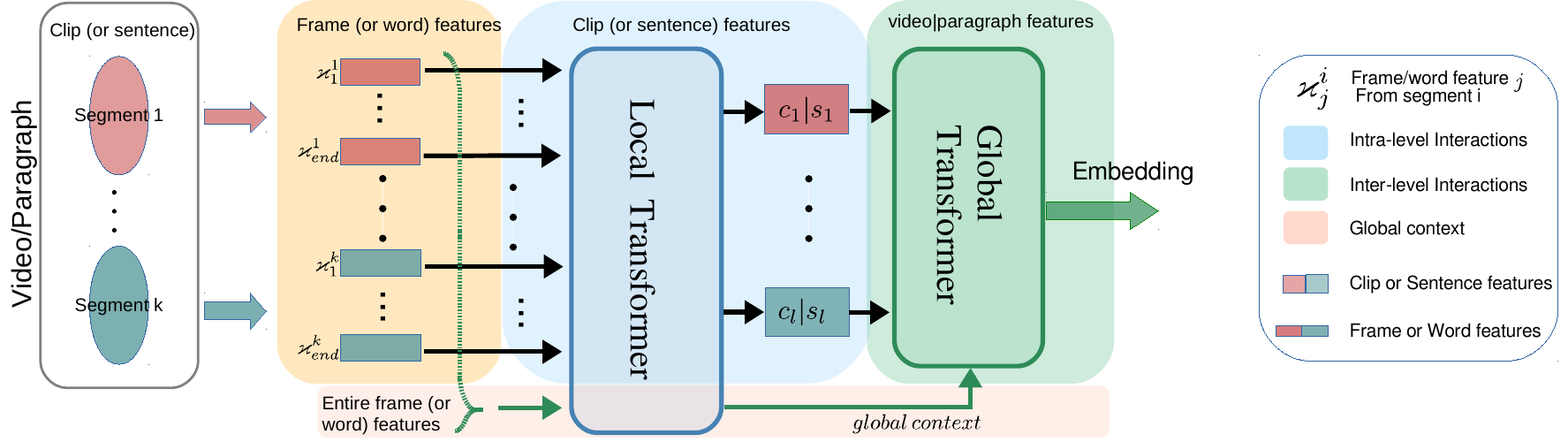}

    \caption{
    \textbf{Overview of architecture} (best viewed in color). We use the same architecture as COOT~\cite{ging2020coot} which consist of two branches: one for video input and one for text input. Since both streams have same design, we here only show one branch. After encoding video data and it's corresponding text, we fed them to local transformer to obtain clip/sentence level features. Then, global transformer aggregates the clip/sentence level features and produces video/paragraph features. The CrossCLR loss is applied to both local (clip/sentence) and global (video/paragraph) features.
    }

    \label{fig:fig_model}
\end{figure}

Note that our architecture is same as COOT. However, We remove all losses used in COOT and only apply CrossCLR in two points. In the architecture~\ref{fig:fig_model}, we apply CrossCLR in two locations: clip and sentence features obtained after local transformer; video and paragraph features obtained after global transformer. In all experiments, we train the model with the following objective:
$L_{local} + 0.6 L_{global}$. For $L_{local}$ we use CrossCLR with queue (hyper-parameters in Table~\ref{tab:hyperparams}) and for $L_{global}$ we use the CrossCLR without queue. Pseudo-code of CrossCLR without queue in Algorithm~\ref{alg:crossclr} is shown in PyTorch style.

    
\section{Hyper-parameters}
We list the hyper-parameters and ranges used during training. We largely follow prior work ~\cite{ging2020coot} for architecture hyper-parameters and for the rest we tuned hyper-parameters based on the performance on validation set. In general, we found retrieval performance to increase with larger queue capacity. In Table~\ref{tab:hyperparams}, we report the hyper-parameters used in the experiments.

\begin{table}[h]
    \caption{
        \textbf{Hyperparameters.} This table shows the hyperparameter ranges we considered and the final choices for
        LSMDC and YouCook2 datasets. First block shows the hyperparameters for optimizer, second block provides the architecture setting and last block shows the hyperparameters for our CrossCLR loss. AF is our Attention-aware Feature Aggregation module.
    }
    \vskip 0.05in
    \centering
    \resizebox{1\columnwidth}{!}{
    \begin{tabular}{l|c|cc}
        \hline
        Hyperparameter                & Range    & LSMDC  & Youcook2 \\ \hline  
        
        Optimizer                           & RAdam                                 & RAdam                  & RAdam     \\
        Learning rate                       & 5e-5  - 1e-3                          & 7e-4                  & 7e-4      \\
        Weight Decay                        & 0                                     & 0                     & 0         \\
        Momentum                            & 0.56                                  & 0.56                   & 0.56      \\
        Warmup Epochs                       & 4                                     & 4                     & 4         \\
        Reduce on Plateau-Patience                        & 6                                     & 6                     & 6         \\
        Reduce on Plateau-Cooldown                        & 4                                     & 4                     & 4         \\ \hline
        Attention Layers                    & 1                                     & 1                     & 1         \\
        Attention Dimension                 & [384, 768]                            & 768                   & 384       \\
        Attention Heads-Local               & [4, 8, 16]                            & 16                     & 8         \\
        Attention Heads-Global              & 1   - 8                               & 8                     & 8         \\
        AF Heads                            & [4, 8]                                & -                     & 2         \\
        Pooler                              & [Max, ATN]                            & Max                     & ATN         \\
        Dropout                             & 1\%   - 10\%                          & 1\%                 & 5\%       \\\hline
        Temperature $\tau$                        & 0.03                                     & 0.03                     & 0.03         \\ 
        Weight scale $\kappa$                        & 1e-5 - 1                                     & 55e-4                     & 35e-4         \\ 
        Intra-modality weight $\lambda$                        & 0.1-1                                     & 65e-1                     & 8e-1         \\ 
        Pruning threshold $\gamma$                        & 0.1-1                                     & 0.9                     & 0.9         \\ 
        Queue size                         & 1,000-10,000                                     & 3,000                     & 5,000         \\

\hline
\end{tabular}
}

\label{tab:hyperparams}
\end{table}


\subsection{Effect of Pruning Threshold}
\begin{figure}[t]
    \centering
    \includegraphics[width=1.0\columnwidth]{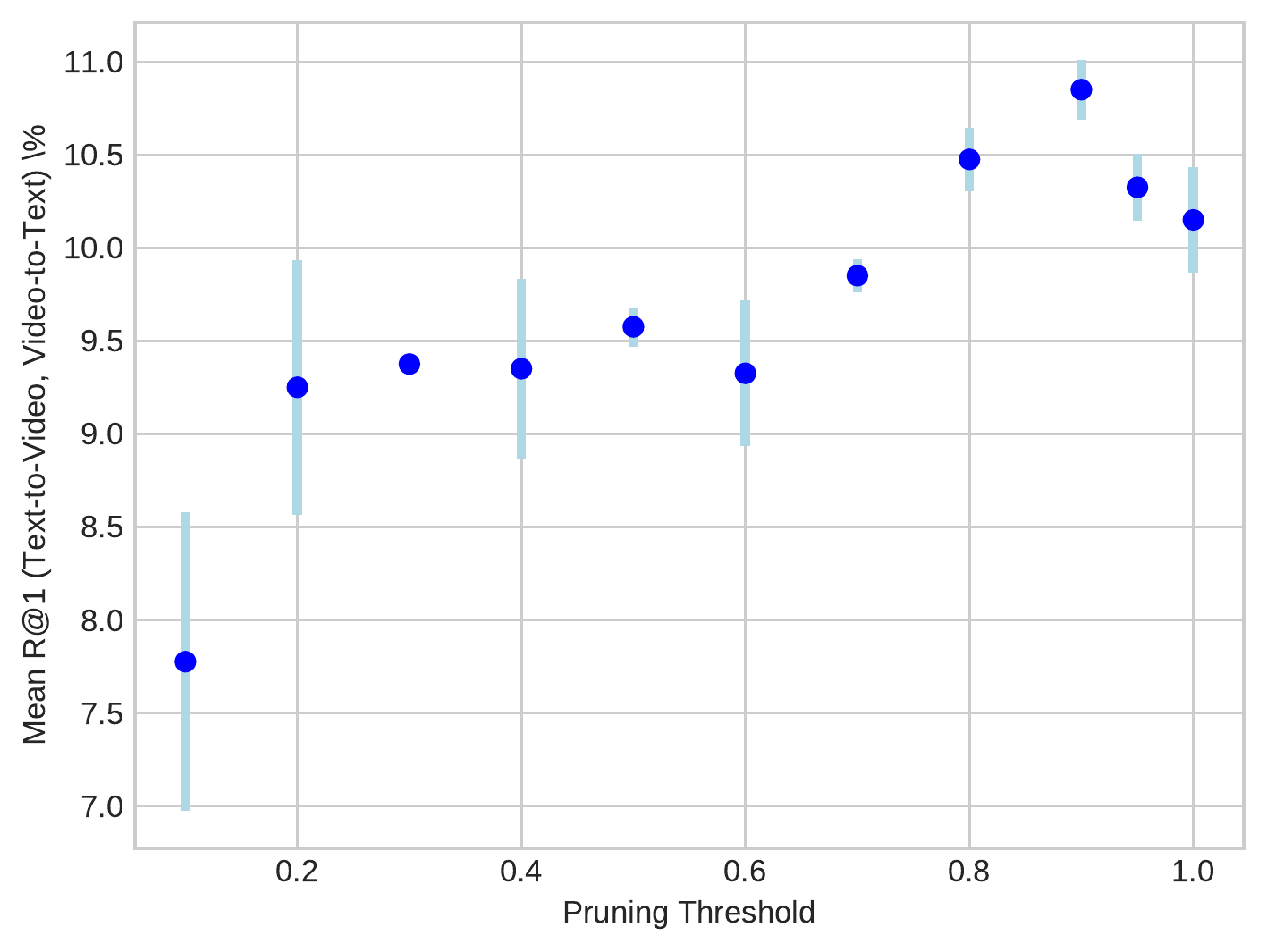}

    \caption{
    \textbf{Impact of Pruning Threshold}. We evaluate the effect of pruning threshold on the retrieval performance for LSMDC dataset using appearnce and action features.   }
    \label{fig:fig_pruning}
\end{figure}

\begin{figure}[t]
    \centering
    \includegraphics[width=1.0\columnwidth]{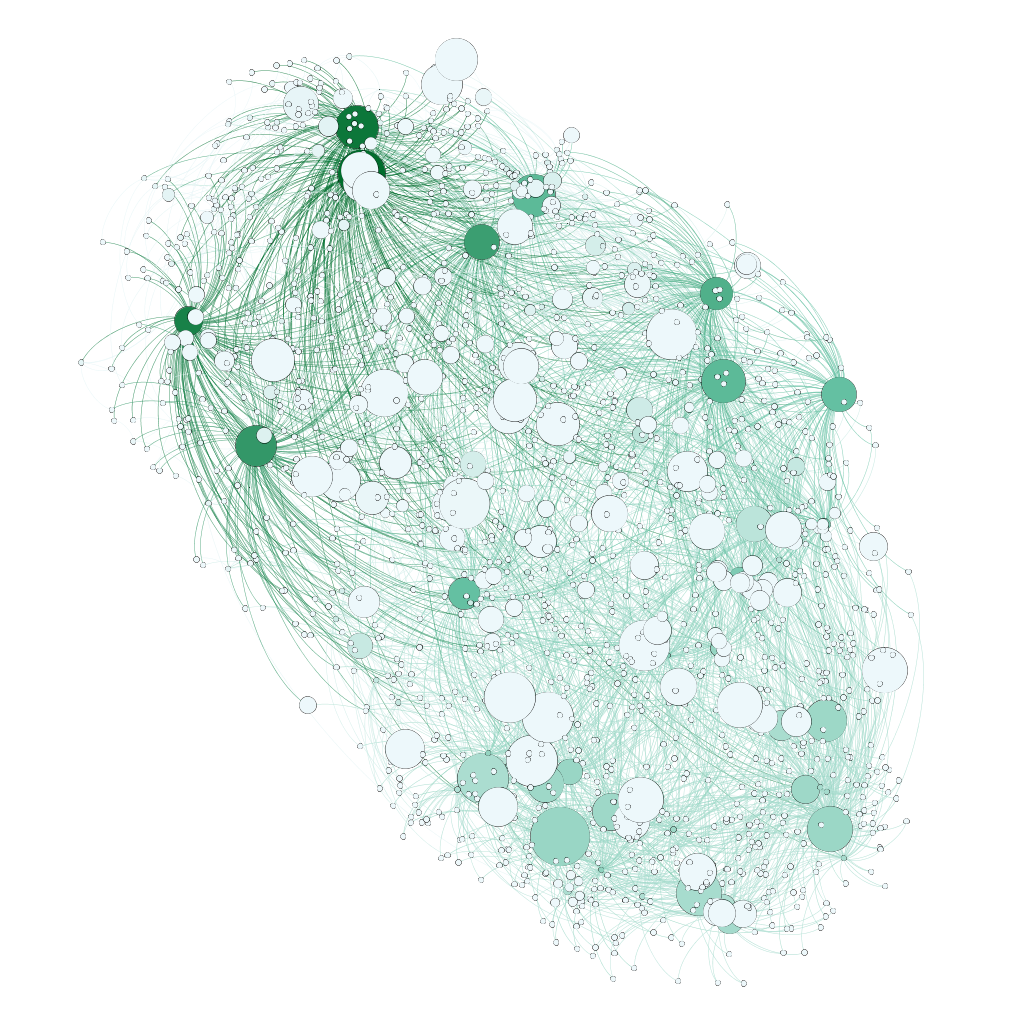}

    \caption{ Graph representation of connections among samples in Youcook2 dataset. Each sample is connected to another sample if their feature similarity is above a certain threshold. Influential samples are densely connected to other samples and therefore share similar features with many other samples. }
    \label{fig:yc2_graph}
\end{figure}

Figure~\ref{fig:yc2_graph}, presents a graph visualization of the Youcook2 dataset embeddings. As can be seen, some samples have dense connections to other samples which means those samples are semantically similar to many samples. Therefore, it's critical to remove these highly similar samples from the negative set to prevent semantic collision, as discussed in section 3.2 of paper.
We change the pruning threshold from $1e1$ to $1$ and show the results in Figure~\ref{fig:fig_pruning}. To prune samples from the negative set, we first divide all scores in each set by the maximum score and then remove the samples which have score above the threshold. Therefore, increasing threshold means removing samples with higher similarity but keeping the other samples which have scores less than threshold. 
We consider thresholds $\gamma \in \{0.1, 0.2, 0.3, 0.4, 0.5, 0.6, 0.7, 0.8, 0.9, 1 \}$ and $\gamma=1$ means no pruning. As shown in Figure~\ref{fig:fig_pruning}, increasing the threshold results in higher performance until $\gamma=0.9$ and after that again performance decreases. We found that for both LSMDC and Youcook2 datasets $\gamma=0.9$ is a reasonable number.

\subsection{Effect of Weight Scale}
\begin{figure}[t]
    \centering
    \includegraphics[width=1.0\columnwidth]{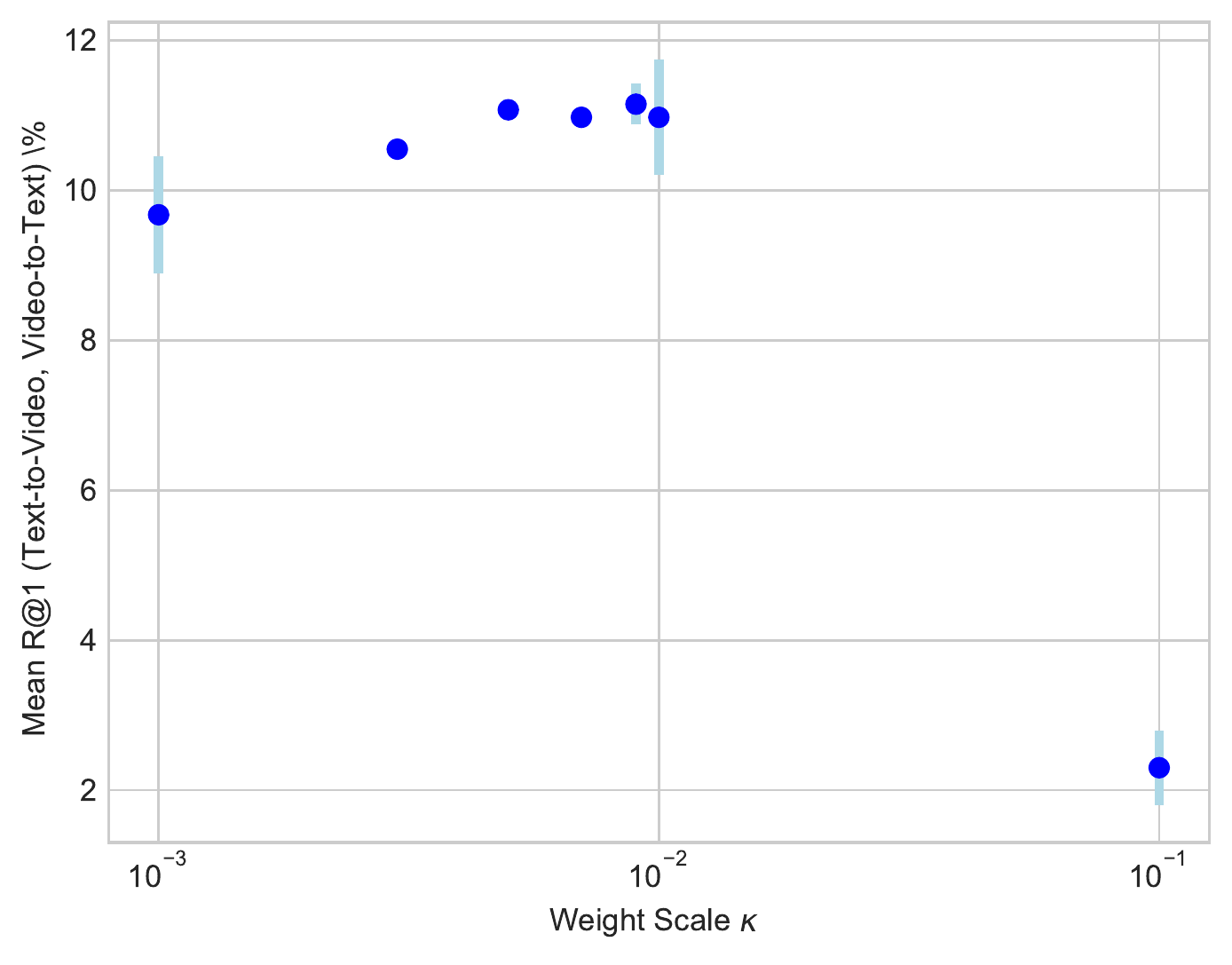}
    \caption{
    \textbf{Impact of Weight Scale $\kappa$}. CrossCLR is stable against small changes in $\kappa$.}
    \label{fig:fig_kappa}
\end{figure}

Figure~\ref{fig:fig_kappa}, shows the retrieval performance with different $\kappa$ values. We train three models for each $\kappa \in \{0.1, 0.01, 0.001, 0.003, 0.005, 0.007, 0.009 \}$. When weight scale is too high, CrossCLR does not converge very well. But with numbers around 0.003-0.005 model performs very well. 
\section{Experiments}

\subsection{Do multiple positives in CrossCLR help?}
In this section, we empirically asses the effect of having multiple positives in objective function. We define our CrossCLR loss with multiple positive samples as following:

\begin{equation}
   \resizebox{.9\hsize}{!}{ 
  $L_x = - \mathbb{E}_{i\in\mathcal M} \left [w(x_i)\log \frac {\delta(x_i, y_i) + \beta\sum\limits_{y_k \in P_y}\delta(x_i, y_k)} {\delta(x_i, y_i) + \sum\limits_{y_k \in \hat{\mathsf{N}}^{E}_{x}} 
  \delta(x_i, y_k) +\lambda \sum\limits_{x_k \in \hat{\mathsf{N}}^{R}_{x}} \delta(x_i, x_k)} \right ] $}
\label{eq_fx}
\end{equation} 
\begin{equation}
   \resizebox{.9\hsize}{!}{ 
  $L_y = - \mathbb{E}_{i\in\mathcal M} \left [w(y_i)\log \frac {\delta(y_i, x_i)+ \beta\sum\limits_{x_k \in P_x}\delta(y_i, x_k)} {\delta(y_i, x_i) + \sum\limits_{x_k \in \hat{\mathsf{N}}^{E}_{y}}
  \delta(y_i, x_k) +\lambda \sum\limits_{y_k \in \hat{\mathsf{N}}^{R}_{y}} \delta(y_i, y_k)} \right ] $}
\label{eq_fy}
\end{equation} 
Where $\beta$ is a scaling factor for extra postive samples, $P_x$ and $P_y$ are positive samples for $L_x$ and $L_y$ respectfully. To construct the positive set, we select the top-K similar samples to the pivot sample among the influential samples. To be precise, $P_x = \{ x_k | k\in topK(\mathcal{I}_y) \}$ and $P_y = \{ y_k | k\in topK(\mathcal{I}_x) \}$. The results are shown in Table~\ref{tab:multiple_positive}. In this experiment, for Youcook2 dataset we found that using $K=2$ and $\beta=0.15$ works best. For LSMDC dataset, we used $K=5$ and $\beta=0.2$. CrossCLR+MP on Youcook2 performs similar to the CrossCLR without multiple positives. However, on LSMDC we observe improvement when multiple positives are used. LSMDC dataset is larger than Youcook2 dataset and therefore it's easier to find very close positive samples to the original sample. Note that, removing influential samples from the negative set reduces the effect of pushing away samples with common semantics. But for multiple positives case, optimization tries to align positive samples and therefore it requires samples to be semantically very close otherwise alignment is wrong.

\begin{table*}[h]
    \caption{\textbf{Effect of Multiple Positives in CrossCLR Loss.} Comparison between our CrossCLR with multiple positives (MP) and standard CrossCLR which does not have multiple positives. For LSMDC experiment, we use appearance and action features together.} 
    \resizebox{2.1\columnwidth}{!}{
    \begin{tabular}{lcccccc||cccccc}
        & \multicolumn{6}{c}{Youcook2} & \multicolumn{6}{c}{LSMDC} \\
        \cmidrule(rl){2-7}  \cmidrule(rl){8-12} 
        & \multicolumn{3}{c}{Text$\implies$Video} & \multicolumn{3}{c}{Video$\implies$Text} & \multicolumn{3}{c}{Text$\implies$Video} & \multicolumn{3}{c}{Video$\implies$Text} \\
         \cmidrule(rl){2-4}  \cmidrule(rl){5-7} \cmidrule(rl){8-10}  \cmidrule(rl){10-12}
          & R@1 & R@5 & R@10  & R@1 & R@5 & R@10 & R@1 & R@5 & R@10  & R@1 & R@5 & R@10  \\ \thickhline
            
            CrossCLR & 19.5\std{0.49} & 45.9\std{0.55} & 58.3\std{0.76} & 18.5\std{0.32} & 44.8\std{0.82} & 57.9\std{0.77}     & 10.9 & 26.2 & 34.7 & 12.0 & 26.1 & 35.3\\
            CrossCLR+MP & 19.3\std{0.51} & 45.3\std{0.85} & 58.2\std{0.77} & 18.4\std{0.45} & 44.4\std{0.61} & 57.6\std{0.81}     & 11.2 & 26.6 & 35.8 & 12.9 & 27.4 & 36.4\\

            \hline

    \end{tabular}
    }
    \label{tab:multiple_positive}
\end{table*}

\subsection{Impact of different modality combinations:}
\begin{table}[t]
    \caption{\textbf{Performance of different modality combinations.} + We feed both modalities to network. $\oplus$ we train the network on each modality separately and then output embeddings are concatenated. }
    \small
    \resizebox{1\columnwidth}{!}{
    \begin{tabular}{lccccc}
        & \multicolumn{5}{c}{Text$\implies$Video} \\
         \cmidrule(rl){2-6}  
          & R@1 & R@5 & R@10  & MdR$\downarrow$  & MnR$\downarrow$\\ \hline
            Random &   0.1 & 0.5 & 1.0 & 500.0 & 500.0  \\ 
            Object &  2.1  & 7.7  & 14.3 & 108.0 & 187.3 \\
            Scene &  5.9  & 18.2  & 24.9 & 57.0 & 132.4 \\
            HowTo100M &  6.4  & 18.9  & 26.4 & 44.0 & 115.0 \\ 
            Appearance &  9.1  & 22.8  & 32.0 & 41.0 & 120.4 \\
            Action &  9.3  & 22.3  & 30.7 & 36.0 & 112.3 \\
            \hline
            Action+Scene &  8.8  & 25.9  & 34.3 & 25.0 & 88.8 \\
            Action+HowTo100M &  9.6  & 24.1  & 35.1 & 26.0 & 86.7 \\
            Action+Object &  9.7  & 23.9  & 31.7 & 31.0 & 102.8 \\
            Action+Appearance &  10.9  & 26.2  & 34.7 & 27.0 & 91.0 \\\hline
            Scene $\oplus$ Object	 &  6.0  & 17.8  & 24.8 & 55.0 & 130.6 \\
            HowTo100M $\oplus$ Object	 &  6.4  & 19.4  & 27.3 & 44.0 & 114.8 \\
            Appearance $\oplus$ Object	 &  8.2  & 22.7  & 29.4 & 44.0 & 121.1 \\
            Scene $\oplus$ HowTo100M	 &  8.7  & 23.6  & 30.1 & 37.0 & 102.7 \\ 
            Action $\oplus$ Object	 &  9.0  & 23.1  & 31.4 & 35.0 & 109.3 \\
            Action $\oplus$ HowTo100M	 &  10.1  & 26.3  & 36.0 & 24.0 & 90.4 \\
            Scene $\oplus$ Appearance	 &  10.2  & 24.2  & 32.0 & 39.0 & 111.0 \\  
            HowTo100M $\oplus$ Appearance	 &  10.5  & 25.5  & 34.4 & 31.0 & 97.3 \\ 
            Action $\oplus$ Scene	 &  11.0  & 24.9  & 34.3 & 30.0 & 96.9 \\
            Action $\oplus$ Appearance	 &  12.0  & 27.7  & 36.2 & 24.0 & 92.4 \\
            \hline
            Scene $\oplus$ Object $\oplus$ HowTo100M         &  8.0  & 22.4  & 30.8 & 37.0 & 106.0 \\
            Object $\oplus$ Appearance $\oplus$ HowTo100M	 &  9.4  & 26.5  & 34.1 & 34.0 & 101.1 \\
            Scene $\oplus$ Appearance $\oplus$ Object	 &  9.2  & 23.3  & 31.9 & 38.0 & 112.4 \\
            Action $\oplus$ Scene $\oplus$ Object         &  10.0  & 25.2  & 33.6 & 32.0 & 100.1 \\
            Scene $\oplus$ Appearance $\oplus$ HowTo100M	 &  10.8  & 26.8  & 34.4 & 32.0 & 97.3 \\
            Action $\oplus$ HowTo100M $\oplus$ Object         &  11.3  & 26.4  & 34.2 & 27.0 & 93.1 \\
            Action $\oplus$ Scene $\oplus$ HowTo100M         &  11.8  & 27.2  & 36.5 & 23.0 & 88.2 \\
            Action $\oplus$ Appearance $\oplus$ Object	 &  12.4  & 27.5  & 35.7 & 28.0 & 95.7 \\
            Action $\oplus$ Appearance $\oplus$ Scene	 &  12.6  & 27.2  & 36.7 & 24.0 & 91.5 \\
            Action $\oplus$ Appearance $\oplus$ HowTo100M	 &  13.4  & 28.4  & 37.8 & 22.0 & 84.9 \\
            \hline

    \end{tabular}
    }
    \label{tab:lsmdc_combine}
\end{table}

We study the importance of different modality experts and their combinations on representation learning in Table~\ref{tab:lsmdc_combine}. Using stronger features result in higher performance. The object expert modality gives the lowest performance and action modality produces the best performance in single modality setting. 
Additionally, we compare the retrieval performance with different combinations of modalities. To combine several modalities, we simply concatenate the features. In addtion, we also trained the network with combining two modalities and inputing them to network (Table~\ref{tab:lsmdc_combine} second block from top shown with $+$). In this paper, we don't study the architecture design for combining multiple modalities and therefore leave this question for future studies.  \\

\begin{figure}[t]
    \centering
    \includegraphics[width=1.0\columnwidth]{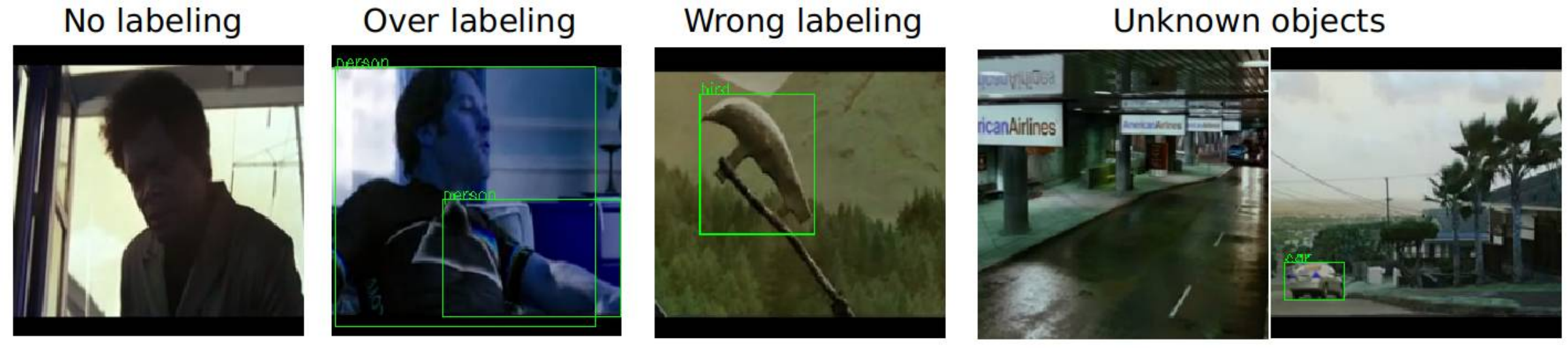}

    \caption{
    \textbf{Challenges in Object Detection for Movie Videos}. We show several qualitative examples of failure in object detection. \tbf{No labeling}: model cannot detect any object in the frame while there is an object, \tbf{Over labeling:} Labeling same object several times, \tbf{Wrong labeling:} Wrong category for the detected object, and \tbf{Unknown objects:} Since some objects are not labeled (or doesn't exist) in the training set of object detection model, these objects that cannot be recognized by the model. }
    \label{fig:qual_object}
\end{figure}

\textbf{Challenges with Object Features:} In Table~\ref{tab:lsmdc_combine}, the object features provide lower performance in comparison to scene or action experts. The reason is that current object detection models are not trained on real-world or large scale movie datasets. Therefore these models cannot detect objects very well due to domain shift or high variations in visual features and high number of objects. In our experiments, we used a Faster RCNN~\cite{ren2016faster} model and used all objects detected with score above 0.7. In Figure~\ref{fig:qual_object}, we visualized some of challenging cases in object detection. 

\begin{algorithm}[t]
    \caption{Pseudocode of CrossCLR (no queue version) in a PyTorch style.}
    \label{alg:code}
    {\fontsize{7.2pt}{0em}\selectfont \texttt{bmm}: batch matrix multiplication; \texttt{mm}: matrix multiplication; \texttt{cat}: concatenation.
    }
    
    \definecolor{codeblue}{rgb}{0.35,0.4,0.7}
    \lstset{
      backgroundcolor=\color{white},
      basicstyle=\fontsize{7.2pt}{7.2pt}\ttfamily\selectfont,
      columns=fullflexible,
      breaklines=true,
      captionpos=b,
      commentstyle=\fontsize{7.2pt}{7.2pt}\color{codeblue},
      keywordstyle=\fontsize{7.2pt}{7.2pt},
    }
    \begin{lstlisting}[language=python]
    # enc_v, enc_t: encoder networks for video and text
    # in_v, in_t: input video and text embeddings (BxD)
    # B: batch size
    # t: temperature, n_keep: 1-(pruning_percentage)
    # t_w: loss weightening temperature
    # mask = 1-np.eye(B)
    # def norm_w(x): return x/sum(x)
    # def mean_w(x,w,t): return sum(x)/(sum(exp(w/t)))
    
    # Encode input video and text embeddings
    emb_v, emb_t = enc_v(in_v), enc_t(in_t)   
 
    # Compute positive and negative logits
    l_vt = mm(emb_v,emb_t.t())/t  #video to text
    l_tv = mm(emb_t,emb_v.t())/t #text to video
    l_vv = (mm(emb_v,emb_v.t())/t)*mask #video to video
    l_tt = (mm(emb_t,emb_t.t()/t)*mask #text to text
 
    # Compute proximity of semantics
    prox_vid = mean(mm(in_v, in_v.t()) * mask) 
    prox_txt = mean(mm(in_t, in_t.t()) * mask) 
    scores_v = prox_vid / max(prox_vid)      
    scores_t = prox_txt / max(prox_txt) 
 
    # Prune samples from intra-modality negative set
    l_vv_p = l_vv[:, scores_v<threshold]   
    l_tt_p = l_tt[:, scores_t<threshold]  
 
    # Prune samples from inter-modality negative set
    l_vt_p = l_vt[:, scores_v<threshold]   
    l_tv_p = l_tv[:, scores_t<threshold] 
    
    # Concatenate positive and negative logits
    l_vtv = cat([l_vt_p, t_w * l_vv_p], dim=1)    
    l_tvt = cat([l_tv_p, t_w * l_tt_p], dim=1)
    
    # compute the loss. Crossentropy without reduction
    labels = arange(l_vt.shape[0])   
    loss_vtv = cross_entropy_loss(l_vtv, labels)
    loss_tvt = cross_entropy_loss(l_tvt, labels)
 
    # Weight losses based on semantic proximity 
    w_vtv = norm_w(prox_vid)
    w_tvt = norm_w(prox_vid)
    loss_vtv = loss_vtv * exp(w_vtv / t_w) 
    loss_tvt = loss_tvt * exp(w_tvt / t_w) 
 
    loss = (mean_w(loss_vtv, w_vtv, t_w) +
              mean_w(loss_tvt, w_tvt, t_w) ) /2
 
    \end{lstlisting}
    \label{alg:crossclr}
 \end{algorithm}

\section{Qualitative Results}
\begin{figure}[t]
    \centering
    \includegraphics[width=1.0\columnwidth]{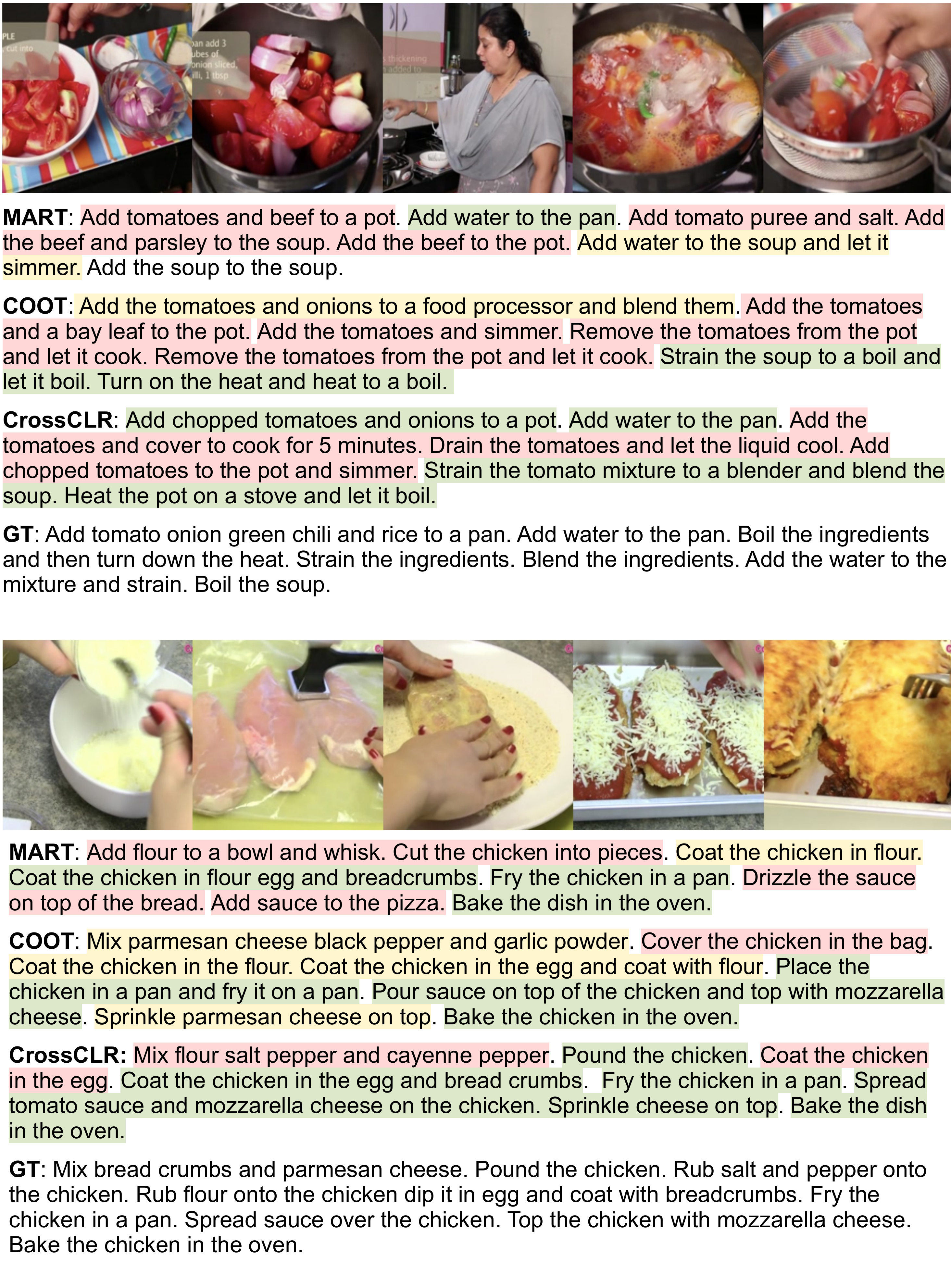}

    \caption{
    \textbf{Captioning samples for Youcook2 dataset}. We randomly selected two samples for video captioning to have a fair comparison with COOT and MART methods. Green: correct captioning, Yellow: Ok but not accurate, and Red: wrong captioning. }
    \label{fig:qual_caption}
\end{figure}

\begin{figure}[t]
    \centering
    \includegraphics[width=1.0\columnwidth]{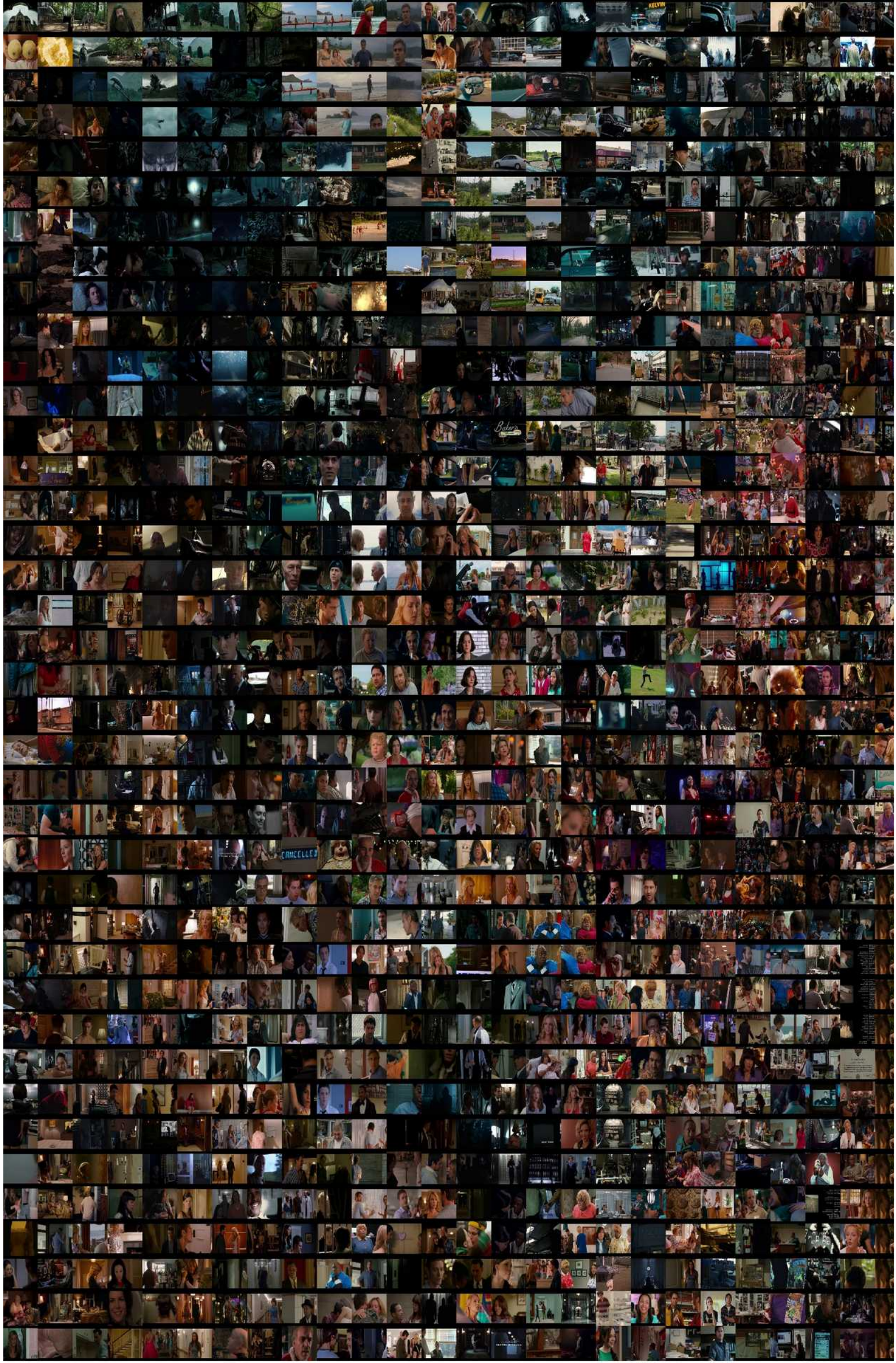}

    \caption{
    \textbf{Visualization of the video frames based on their text embeddings}. We use t-SNE to project text embeddings to 2D space and then present each point with it's corresponding video frame. }

    \label{fig:qual_tsne}
\end{figure}

Figure~\ref{fig:qual_caption}, shows two qualitative examples for video captioning on Youcook2 dataset (samples are selected randomly for a fair comparison with COOT and MART).
In Figure~\ref{fig:qual_tsne}, we visualize the video frames in the embedding space based on distances between text embeddings for Youcook2 dataset. Two videos are close to each other if their text features are similar to each other. We use t-SNE and project the text embeddings to 2D space and then visualize each point with it's corresponding video frame. As can be seen, the text embeddings are clustered semantically and aligned with visual semantics.

\end{appendices}

{\small
\bibliographystyle{ieee_fullname}
\bibliography{paper_for_arxiv}

\begin{thebibliography}{10}\itemsep=-1pt

\bibitem{arora2019theoretical}
Sanjeev Arora, Hrishikesh Khandeparkar, Mikhail Khodak, Orestis Plevrakis, and
  Nikunj Saunshi.
\newblock A theoretical analysis of contrastive unsupervised representation
  learning, 2019.

\bibitem{caron_nips2020_swav}
Mathilde Caron, Ishan Misra, Julien Mairal, Priya Goyal, Piotr Bojanowski, and
  Armand Joulin.
\newblock {Unsupervised Learning of Visual Features by Contrasting Cluster
  Assignments}.
\newblock {\em arXiv preprint arXiv:2006.09882}, 2020.

\bibitem{chen_icml2020_simclr}
Ting Chen, Simon Kornblith, Mohammad Norouzi, and Geoffrey Hinton.
\newblock {A Simple Framework for Contrastive Learning of Visual
  Representations}.
\newblock In {\em ICML}. 2020.

\bibitem{chen2020simple}
Ting Chen, Simon Kornblith, Mohammad Norouzi, and Geoffrey Hinton.
\newblock A simple framework for contrastive learning of visual
  representations.
\newblock In {\em International conference on machine learning}, pages
  1597--1607. PMLR, 2020.

\bibitem{chen_arxiv2020_simclrv2}
Ting Chen, Simon Kornblith, Kevin Swersky, Mohammad Norouzi, and Geoffrey
  Hinton.
\newblock {Big Self-Supervised Models are Strong Semi-Supervised Learners}.
\newblock {\em arXiv preprint arXiv:2006.10029}, 2020.

\bibitem{chen2020improved}
Xinlei Chen, Haoqi Fan, Ross Girshick, and Kaiming He.
\newblock Improved baselines with momentum contrastive learning.
\newblock {\em arXiv preprint arXiv:2003.04297}, 2020.

\bibitem{chuang2020debiased}
Ching-Yao Chuang, Joshua Robinson, Yen-Chen Lin, Antonio Torralba, and Stefanie
  Jegelka.
\newblock Debiased contrastive learning.
\newblock {\em Advances in Neural Information Processing Systems}, 33, 2020.

\bibitem{dai2019transformerxl}
Zihang Dai, Zhilin Yang, Yiming Yang, Jaime Carbonell, Quoc~V. Le, and Ruslan
  Salakhutdinov.
\newblock Transformer-xl: Attentive language models beyond a fixed-length
  context, 2019.

\bibitem{denkowski-lavie-2014-meteor}
Michael Denkowski and Alon Lavie.
\newblock Meteor universal: Language specific translation evaluation for any
  target language.
\newblock In {\em Proceedings of the Ninth Workshop on Statistical Machine
  Translation}, pages 376--380, Baltimore, Maryland, USA, Jun 2014. Association
  for Computational Linguistics.

\bibitem{DongLXJH0W19}
Jianfeng Dong, Xirong Li, Chaoxi Xu, Shouling Ji, Yuan He, Gang Yang, and Xun
  Wang.
\newblock Dual encoding for zero-example video retrieval.
\newblock In {\em {IEEE} Conference on Computer Vision and Pattern Recognition,
  {CVPR} 2019, Long Beach, CA, USA, June 16-20, 2019}, pages 9346--9355.
  Computer Vision Foundation / {IEEE}, 2019.

\bibitem{gabeur2020multimodal}
Valentin Gabeur, Chen Sun, Karteek Alahari, and Cordelia Schmid.
\newblock Multi-modal transformer for video retrieval, 2020.

\bibitem{ging2020coot}
Simon Ging, Mohammadreza Zolfaghari, Hamed Pirsiavash, and Thomas Brox.
\newblock Coot: Cooperative hierarchical transformer for video-text
  representation learning.
\newblock In {\em Advances on Neural Information Processing Systems (NeurIPS)},
  2020.

\bibitem{grill_arxiv2020_byol}
Jean-Bastien Grill, Florian Strub, Florent Altché, Corentin Tallec, Pierre~H.
  Richemond, Elena Buchatskaya, Carl Doersch, Bernardo~Avila Pires,
  Zhaohan~Daniel Guo, Mohammad~Gheshlaghi Azar, Bilal Piot, Koray Kavukcuoglu,
  Rémi Munos, and Michal Valko.
\newblock {Bootstrap Your Own Latent: A New Approach to Self-Supervised
  Learning}.
\newblock {\em arXiv preprint arXiv:2006.07733}, 2020.

\bibitem{1640964}
R. {Hadsell}, S. {Chopra}, and Y. {LeCun}.
\newblock Dimensionality reduction by learning an invariant mapping.
\newblock In {\em 2006 IEEE Computer Society Conference on Computer Vision and
  Pattern Recognition (CVPR'06)}, volume~2, pages 1735--1742, 2006.

\bibitem{Han20}
Tengda Han, Weidi Xie, and Andrew Zisserman.
\newblock Self-supervised co-training for video representation learning.
\newblock In {\em Neurips}, 2020.

\bibitem{he_cvpr2020_moco}
Kaiming He, Haoqi Fan, Yuxin Wu, Saining Xie, and Ross Girshick.
\newblock {Momentum Contrast for Unsupervised Visual Representation Learning}.
\newblock In {\em CVPR}. 2020.

\bibitem{hessel_caption19}
Jack Hessel, Bo Pang, Zhenhai Zhu, and Radu Soricut.
\newblock A case study on combining {ASR} and visual features for generating
  instructional video captions.
\newblock {\em CoRR}, abs/1910.02930, 2019.

\bibitem{triplet3}
Elad Hoffer and Nir Ailon.
\newblock Deep metric learning using triplet network.
\newblock In Aasa Feragen, Marcello Pelillo, and Marco Loog, editors, {\em
  Similarity-Based Pattern Recognition - Third International Workshop, {SIMBAD}
  2015, Copenhagen, Denmark, October 12-14, 2015, Proceedings}, volume 9370 of
  {\em Lecture Notes in Computer Science}, pages 84--92. Springer, 2015.

\bibitem{kalantidis_nips2020_hardneg}
Yannis Kalantidis, Mert~Bulent Sariyildiz, Noe Pion, Philippe Weinzaepfel, and
  Diane Larlus.
\newblock {Hard Negative Mixing for Contrastive Learning}.
\newblock {\em NeurIPS}, 2020.

\bibitem{khosla2020supervised}
Prannay Khosla, Piotr Teterwak, Chen Wang, Aaron Sarna, Yonglong Tian, Phillip
  Isola, Aaron Maschinot, Ce Liu, and Dilip Krishnan.
\newblock Supervised contrastive learning.
\newblock {\em arXiv preprint arXiv:2004.11362}, 2020.

\bibitem{klein2015associating}
Benjamin Klein, Guy Lev, Gil Sadeh, and Lior Wolf.
\newblock Associating neural word embeddings with deep image representations
  using fisher vectors.
\newblock In {\em Proceedings of the IEEE Conference on Computer Vision and
  Pattern Recognition}, pages 4437--4446, 2015.

\bibitem{mart}
Jie Lei, Liwei Wang, Yelong Shen, Dong Yu, Tamara~L Berg, and Mohit Bansal.
\newblock Mart: Memory-augmented recurrent transformer for coherent video
  paragraph captioning.
\newblock In {\em ACL}, 2020.

\bibitem{lin-2004-rouge}
Chin-Yew Lin.
\newblock Rouge: A package for automatic evaluation of summaries.
\newblock In {\em Text Summarization Branches Out}, pages 74--81, Barcelona,
  Spain, Jul 2004. Association for Computational Linguistics.

\bibitem{LiuANZ19}
Yang Liu, Samuel Albanie, Arsha Nagrani, and Andrew Zisserman.
\newblock Use what you have: Video retrieval using representations from
  collaborative experts.
\newblock In {\em 30th British Machine Vision Conference 2019, {BMVC} 2019,
  Cardiff, UK, September 9-12, 2019}, page 279. {BMVA} Press, 2019.

\bibitem{ma2020learning}
Shuang Ma, Zhaoyang Zeng, Daniel McDuff, and Yale Song.
\newblock Learning audio-visual representations with active contrastive coding,
  2020.

\bibitem{miech19endtoend}
Antoine Miech, Jean-Baptiste Alayrac, Lucas Smaira, Ivan Laptev, Josef Sivic,
  and Andrew Zisserman.
\newblock {E}nd-to-{E}nd {L}earning of {V}isual {R}epresentations from
  {U}ncurated {I}nstructional {V}ideos.
\newblock In {\em CVPR}, 2020.

\bibitem{miech2018learning}
Antoine Miech, Ivan Laptev, and Josef Sivic.
\newblock Learning a text-video embedding from incomplete and heterogeneous
  data.
\newblock {\em arXiv preprint arXiv:1804.02516}, 2018.

\bibitem{miech19howto100m}
Antoine Miech, Dimitri Zhukov, Jean-Baptiste Alayrac, Makarand Tapaswi, Ivan
  Laptev, and Josef Sivic.
\newblock How{T}o100{M}: {L}earning a {T}ext-{V}ideo {E}mbedding by {W}atching
  {H}undred {M}illion {N}arrated {V}ideo {C}lips.
\newblock In {\em ICCV}, 2019.

\bibitem{mithun2018learning}
Niluthpol~C Mithun, Juncheng Li, Florian Metze, and Amit~K Roy-Chowdhury.
\newblock Learning joint embedding with multimodal cues for cross-modal
  video-text retrieval.
\newblock In {\em ICMR}. ACM, 2018.

\bibitem{morgado2020learning}
Pedro Morgado, Yi Li, and Nuno Vasconcelos.
\newblock Learning representations from audio-visual spatial alignment, 2020.

\bibitem{papineni-etal-2002-bleu}
Kishore Papineni, Salim Roukos, Todd Ward, and Wei-Jing Zhu.
\newblock Bleu: a method for automatic evaluation of machine translation.
\newblock In {\em Proceedings of the 40th Annual Meeting of the Association for
  Computational Linguistics}, pages 311--318, Philadelphia, Pennsylvania, USA,
  Jul 2002. Association for Computational Linguistics.

\bibitem{portilloquintero2021straightforward}
Jesús~Andrés Portillo-Quintero, José~Carlos Ortiz-Bayliss, and Hugo
  Terashima-Marín.
\newblock A straightforward framework for video retrieval using clip, 2021.

\bibitem{rui2020}
Rui Qian, Tianjian Meng, Boqing Gong, Ming{-}Hsuan Yang, Huisheng Wang,
  Serge~J. Belongie, and Yin Cui.
\newblock Spatiotemporal contrastive video representation learning.
\newblock {\em CoRR}, abs/2008.03800, 2020.

\bibitem{radford2021learning}
Alec Radford, Jong~Wook Kim, Chris Hallacy, Aditya Ramesh, Gabriel Goh,
  Sandhini Agarwal, Girish Sastry, Amanda Askell, Pamela Mishkin, Jack Clark,
  et~al.
\newblock Learning transferable visual models from natural language
  supervision.
\newblock {\em Image}, 2:T2, 2021.

\bibitem{ren2016faster}
Shaoqing Ren, Kaiming He, Ross Girshick, and Jian Sun.
\newblock Faster r-cnn: Towards real-time object detection with region proposal
  networks, 2016.

\bibitem{robinson_iclr2021_hardneg}
Joshua Robinson, Ching-Yao Chuang, Suvrit Sra, and Stefanie Jegelka.
\newblock {Contrastive Learning with Hard Negative Samples}.
\newblock {\em ICLR}, 2021.

\bibitem{7298940}
A. {Rohrbach}, M. {Rohrbach}, N. {Tandon}, and B. {Schiele}.
\newblock A dataset for movie description.
\newblock In {\em 2015 IEEE Conference on Computer Vision and Pattern
  Recognition (CVPR)}, pages 3202--3212, 2015.

\bibitem{zhang_arxiv2020_adaclr}
Xiaokang~Yang Shaofeng~Zhang, Junchi~Yan.
\newblock {Self-supervised representation learning via adaptive hard-positive
  mining}.
\newblock {\em https://openreview.net/forum?id=aLIbnLY9NtH}, 2020.

\bibitem{sohn2020learning}
Kihyuk Sohn, Chun-Liang Li, Jinsung Yoon, Minho Jin, and Tomas Pfister.
\newblock Learning and evaluating representations for deep one-class
  classification, 2020.

\bibitem{8953619}
Y. {Suh}, B. {Han}, W. {Kim}, and K.~M. {Lee}.
\newblock Stochastic class-based hard example mining for deep metric learning.
\newblock In {\em 2019 IEEE/CVF Conference on Computer Vision and Pattern
  Recognition (CVPR)}, pages 7244--7252, 2019.

\bibitem{9009570}
C. {Sun}, A. {Myers}, C. {Vondrick}, K. {Murphy}, and C. {Schmid}.
\newblock Videobert: A joint model for video and language representation
  learning.
\newblock In {\em 2019 IEEE/CVF International Conference on Computer Vision
  (ICCV)}, pages 7463--7472, 2019.

\bibitem{sun2019videobert}
Chen Sun, Austin Myers, Carl Vondrick, Kevin Murphy, and Cordelia Schmid.
\newblock Videobert: A joint model for video and language representation
  learning.
\newblock In {\em Proceedings of the IEEE International Conference on Computer
  Vision}, pages 7464--7473, 2019.

\bibitem{tran2019video}
Du Tran, Heng Wang, Lorenzo Torresani, and Matt Feiszli.
\newblock Video classification with channel-separated convolutional networks.
\newblock In {\em Proceedings of the IEEE/CVF International Conference on
  Computer Vision}, pages 5552--5561, 2019.

\bibitem{abs-1807-03748}
A{\"{a}}ron van~den Oord, Yazhe Li, and Oriol Vinyals.
\newblock Representation learning with contrastive predictive coding.
\newblock {\em CoRR}, abs/1807.03748, 2018.

\bibitem{cider}
R. Vedantam, C.~L. Zitnick, and D. Parikh.
\newblock Cider: Consensus-based image description evaluation.
\newblock In {\em 2015 IEEE Conference on Computer Vision and Pattern
  Recognition (CVPR)}, pages 4566--4575, 2015.

\bibitem{triplet1}
Jiang Wang, Yang Song, Thomas Leung, Chuck Rosenberg, Jingbin Wang, James
  Philbin, Bo Chen, and Ying Wu.
\newblock Learning fine-grained image similarity with deep ranking.
\newblock {\em CoRR}, abs/1404.4661, 2014.

\bibitem{triplet2}
Liwei Wang, Yin Li, and Svetlana Lazebnik.
\newblock Learning deep structure-preserving image-text embeddings.
\newblock {\em CoRR}, abs/1511.06078, 2015.

\bibitem{Xinshao19}
Xinshao Wang, Elyor Kodirov, Yang Hua, and Neil~Martin Robertson.
\newblock Instance cross entropy for deep metric learning.
\newblock {\em CoRR}, abs/1911.09976, 2019.

\bibitem{8578491}
Z. {Wu}, Y. {Xiong}, S.~X. {Yu}, and D. {Lin}.
\newblock Unsupervised feature learning via non-parametric instance
  discrimination.
\newblock In {\em 2018 IEEE/CVF Conference on Computer Vision and Pattern
  Recognition}, pages 3733--3742, 2018.

\bibitem{ngramrepetition}
Yilei Xiong, Bo Dai, and Dahua Lin.
\newblock Move forward and tell: A progressive generator of video descriptions.
\newblock In {\em ECCV}, 2018.

\bibitem{jsf18}
Youngjae Yu, Jongseok Kim, and Gunhee Kim.
\newblock A joint sequence fusion model for video question answering and
  retrieval.
\newblock {\em CoRR}, abs/1808.02559, 2018.

\bibitem{yu2017endtoend}
Youngjae Yu, Hyungjin Ko, Jongwook Choi, and Gunhee Kim.
\newblock End-to-end concept word detection for video captioning, retrieval,
  and question answering, 2017.

\bibitem{zhang2020resnest}
Hang Zhang, Chongruo Wu, Zhongyue Zhang, Yi Zhu, Zhi Zhang, Haibin Lin, Yue
  Sun, Tong He, Jonas Mueller, R Manmatha, et~al.
\newblock Resnest: Split-attention networks.
\newblock {\em arXiv preprint arXiv:2004.08955}, 2020.

\bibitem{zhou2017places}
Bolei Zhou, Agata Lapedriza, Aditya Khosla, Aude Oliva, and Antonio Torralba.
\newblock Places: A 10 million image database for scene recognition.
\newblock {\em IEEE Transactions on Pattern Analysis and Machine Intelligence},
  2017.

\bibitem{ZhXuCoCVPR18}
Luowei Zhou, Chenliang Xu, and Jason~J Corso.
\newblock Towards automatic learning of procedures from web instructional
  videos.
\newblock In {\em AAAI Conference on Artificial Intelligence}, 2018.

\bibitem{zhou2018endtoend}
Luowei Zhou, Yingbo Zhou, Jason~J. Corso, Richard Socher, and Caiming Xiong.
\newblock End-to-end dense video captioning with masked transformer, 2018.

\bibitem{actbert20}
L. {Zhu} and Y. {Yang}.
\newblock Actbert: Learning global-local video-text representations.
\newblock In {\em 2020 IEEE/CVF Conference on Computer Vision and Pattern
  Recognition (CVPR)}, pages 8743--8752, 2020.

\end{thebibliography}
}

\end{document}